%% file: main.tex
\documentclass[letterpaper]{article} 
\usepackage{aaai25_arxiv}  
\usepackage{times}  
\usepackage{helvet}  
\usepackage{courier}  
\usepackage[hyphens]{url}  
\usepackage{graphicx} 
\usepackage{natbib}  
\usepackage{caption} 
\usepackage{algorithm}
\usepackage{algorithmic}

\usepackage{newfloat}
\usepackage{listings}
\DeclareCaptionStyle{ruled}{labelfont=normalfont,labelsep=colon,strut=off} 
\floatstyle{ruled}
\newfloat{listing}{tb}{lst}{}
\floatname{listing}{Listing}
\usepackage{hyperref} 
\usepackage{booktabs}       
\usepackage{bm}
\usepackage{amsmath}
\usepackage{amssymb}
\usepackage{cleveref}
\usepackage{multirow}
\usepackage{enumitem}
\usepackage{subcaption}

\input{command}

\title{Efficient Self-Supervised Video Hashing with Selective State Spaces}
\author{
    Jinpeng Wang\textsuperscript{\rm 1}\thanks{The first three authors contributed equally to this work.}, 
    Niu Lian\textsuperscript{\rm 2*}, 
    Jun Li\textsuperscript{\rm 2*}, 
    Yuting Wang\textsuperscript{\rm 1}, 
    Yan Feng\textsuperscript{\rm 4}, \\
    Bin Chen\textsuperscript{\rm 2,3}\thanks{Corresponding author.}, 
    Yongbing Zhang\textsuperscript{\rm 2}, 
    Shu-Tao Xia\textsuperscript{\rm 1,3}
}
\affiliations{
    \textsuperscript{\rm 1}Tsinghua Shenzhen International Graduate School, Tsinghua University\\
    \textsuperscript{\rm 2}Harbin Institute of Technology, Shenzhen\\ 
    \textsuperscript{\rm 3}Research Center of Artificial Intelligence, Peng Cheng Laboratory \\
    \textsuperscript{\rm 4}Meituan, Beijing \\
    wjp20@mails.tsinghua.edu.cn,
    \{220110904,220110924\}@stu.hit.edu.cn,
    huangmozhi9527@gmail.com,
    fengyan14@meituan.com, 
    chenbin2021@hit.edu.cn,
    ybzhang08@hit.edu.cn,
    xiast@sz.tsinghua.edu.cn
}

\begin{document}

\maketitle
\input{sections/Abstract}
\input{sections/Introduction}
\input{sections/RelatedWork}
\input{sections/Method}
\input{sections/Experiments}
\input{sections/Conclusions}

\bibliography{main}

\end{document}

%% file: command.tex
\newcommand{\ie}{\emph{i.e.},~}

\newcommand{\wrt}{\emph{w.r.t.}~}

\renewcommand{\paragraph}[1]{\medskip\noindent\textbf{#1.~}}

\newcommand{\dak}[1]{\left\{#1\right\}}

\newcommand{\shuk}[1]{\left\lVert#1\right\rVert}

\newcommand{\opseq}[3]{{#1_1 #3 #1_2 #3 \cdots #3 #1_{#2}}}

\newcommand{\bmb}{\bm{b}}

\newcommand{\bmh}{\bm{h}}

\newcommand{\bmx}{\bm{x}}
\newcommand{\bmy}{\bm{y}}

\newcommand{\bmone}{\bm{1}}

\newcommand{\bmtheta}{\bm{\theta}}

\newcommand{\bmphi}{\bm{\phi}}

\newcommand{\bmA}{\bm{A}}
\newcommand{\bmB}{\bm{B}}
\newcommand{\bmC}{\bm{C}}

\newcommand{\bmE}{\bm{E}}
\newcommand{\bmF}{\bm{F}}
\newcommand{\bmG}{\bm{G}}
\newcommand{\bmH}{\bm{H}}
\newcommand{\bmI}{\bm{I}}

\newcommand{\bmS}{\bm{S}}

\newcommand{\bmW}{\bm{W}}

\newcommand{\bmTheta}{\bm{\Theta}}

\newcommand{\bmUpsilon}{\bm{\Upsilon}}
\newcommand{\bmPhi}{\bm{\Phi}}
\newcommand{\bmPsi}{\bm{\Psi}}

\newcommand{\calC}{\mathcal{C}}
\newcommand{\calD}{\mathcal{D}}
\newcommand{\calE}{\mathcal{E}}

\newcommand{\calL}{\mathcal{L}}
\newcommand{\calM}{\mathcal{M}}

\newcommand{\calS}{\mathcal{S}}

\newcommand{\bbR}{\mathbb{R}}

\newcommand{\modelname}{S5VH}

%% file: sections/Abstract.tex
\begin{abstract}
Self-supervised video hashing (SSVH) is a practical task in video indexing and retrieval. 
Although Transformers are predominant in SSVH for their impressive temporal modeling capabilities, they often suffer from computational and memory inefficiencies. 
Drawing inspiration from Mamba, an advanced state-space model, we explore its potential in SSVH to achieve a better balance between efficacy and efficiency. 
We introduce \modelname{}, a Mamba-based video hashing model with an improved self-supervised learning paradigm. 
Specifically, we design bidirectional Mamba layers for both the encoder and decoder, which are effective and efficient in capturing temporal relationships thanks to the data-dependent selective scanning mechanism with linear complexity. 
In our learning strategy, we transform global semantics in the feature space into semantically consistent and discriminative hash centers, followed by a center alignment loss as a global learning signal. 
Our self-local-global (SLG) paradigm significantly improves learning efficiency, leading to faster and better convergence. 
Extensive experiments demonstrate \modelname{}'s  improvements over state-of-the-art methods, superior transferability, and scalable advantages in inference efficiency. 
\end{abstract}

\begin{links}
    \link{Code}{https://github.com/gimpong/AAAI25-S5VH}
\end{links}

%% file: sections/Introduction.tex
\section{Introduction}
\label{sec: introduction}

\begin{figure}[t]
    \centering
    \includegraphics[width=\linewidth]{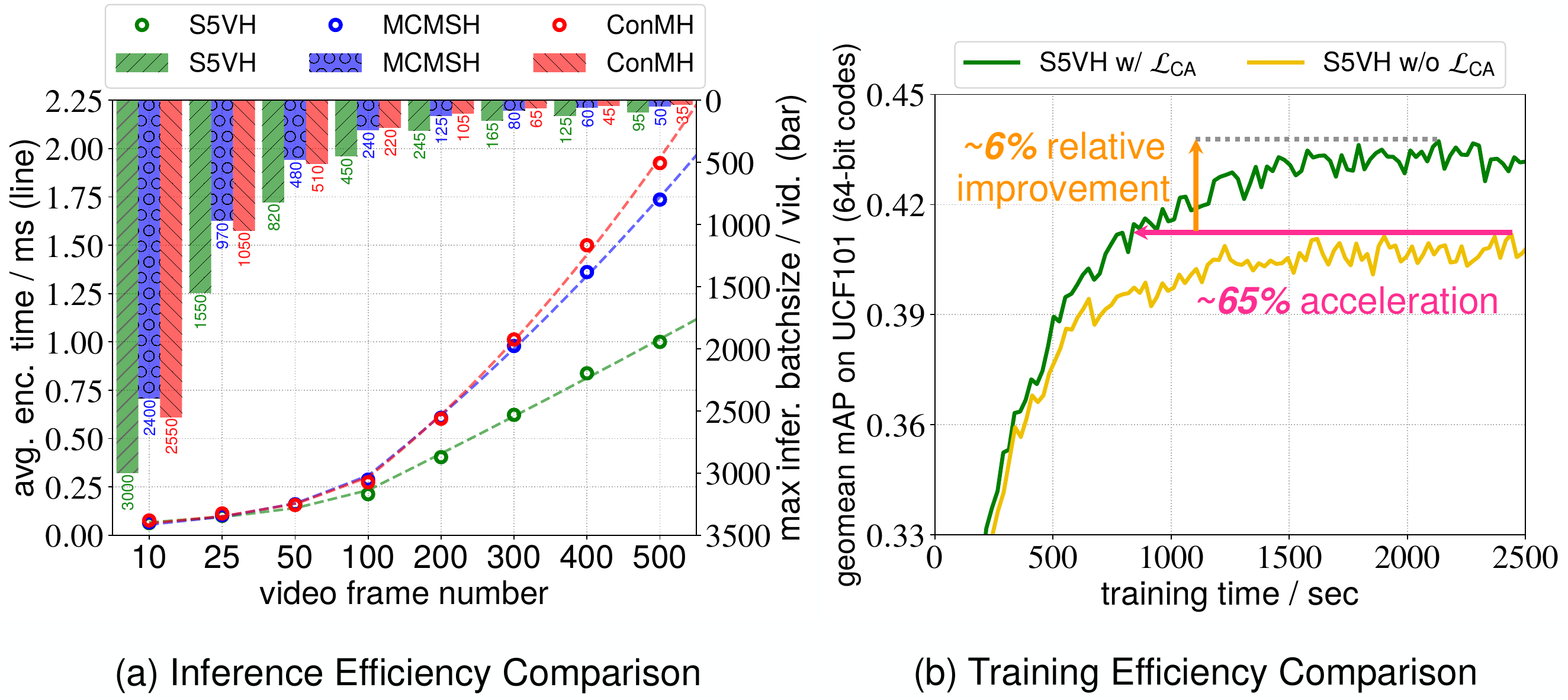}
    \caption{Highlights: 
    (\textbf{a}) Our \modelname{} based on Mamba exhibits lower inference overheads on memory and computation. The efficiency advantages are scalable and more notable under larger frame numbers.
    (\textbf{b}) The introduced global learning signal in the hash space effectively enhances
    training efficiency, showing faster and better convergence.}
    \label{fig:intro}
\end{figure}

Content-based retrieval is a basic component in video search and recommendation. 
Hashing has been widely explored in this context to facilitate fast retrieval and reduce memory footprint \cite{gao2023backdoor,sun2023hierarchical,sun2024dual,HuggingHash2_2024}. 
Video hashing has evolved from traditional methods with handcrafted features to advanced deep approaches with substantially improved retrieval performance, where self-supervised video hashing (SSVH) has gained increasing attention, given the ubiquity of large-scale unlabeled video data and evoked by the rapid progress of self-supervised learning in recent years \cite{MoCo_2020}.

In SSVH, temporal modeling is essential for video understanding and hash code learning. 
Early approaches \cite{SSVH_2018,UVVH_2019} utilized nonlinear RNNs to process frame features, which suffered from gradient vanishing (or exploding) and struggled over long-range dependencies. 
In contrast, Transformer-based models \cite{BTH_2021,ConMH_2023} were shown to capture temporal semantics better, thanks to self-attentive interactions across all frames. 
Despite state-of-the-art retrieval performance, the quadratically growing complexity of Transformers in computation and memory to the frame number renders scalability limitations. 
Pursuing the optimal trade-off between efficacy and efficiency remains an under-explored problem. 

Recent advances have sparked interest in state space models (SSMs) \cite{LSSL_2021, DSS_2022}, where we find Mamba \cite{Mamba_2023}, an improved variant of structured SSMs (S4) \cite{S4_2021}, can bring insightful solutions to the above problem. 
In detail, the data-dependent selective mechanism in Mamba ensures focusing on essential information while filtering out irrelevant noises regarding the input, which helps to capture temporal dynamics and understand long-range relations in videos. 
Moreover, Mamba's linear complexity regarding sequence length promises superior scalability for video processing, and its GPU-friendly implementation also aligns with SSVH's pursuit of efficient inference.
These advantages motivate us to explore Mamba's potential for SSVH. 

In this paper, we introduce \textbf{\modelname{}}\footnote{{\scriptsize Abbr. of \underline{\textbf{S}}elf-\underline{\textbf{S}}upervised \underline{\textbf{S}}elective \underline{\textbf{S}}tate-\underline{\textbf{S}}pace \underline{\textbf{V}}ideo \underline{\textbf{H}}ashing (\textbf{SSSSSVH})}.}, the first selective state-space approach for SSVH. 
For \emph{network design}, we explore a \textbf{Mamba-based hashing network}, where each encoder or decoder layer comprises a bidirectional Mamba module for effective and efficient temporal modeling. 
For \emph{self-supervised hash learning}, we develop a \textbf{self-local-global (SLG) paradigm} that excavates hierarchical learning signals to enhance training efficiency. 
Our motivation stems from the Mamba's serial nature that restricts training throughout \cite{VMamba_2024}, and we resort to a practical and hashing-oriented solution by maximizing sample efficiency. 

State-of-the-art approaches \cite{ConMH_2023,CHAIN_2023,TSVH_2024} typically follow the \emph{intra}- and \emph{inter}-sample learning paradigm, where intra-sample (\ie \emph{self}) signals refer to recovery tasks with various data augmentations for video understanding, while the inter-sample signals refer to contrastive tasks between videos to obtain discriminative hash codes. 
We note that inter-sample signals are subject to \emph{local} signals reflected by individual samples, whose efficiency is limited by the (negative) sampling. 
We further integrate a \emph{global} signal in the Hamming space to complement the learning paradigm. 
We start with clustering video feature space to summarize global semantic structure by cluster-level similarity. 
Then, we introduce a novel hash center generation algorithm to transform the global semantics into well-separated and semantically consistent hash centers. 
Based on them, we propose a center alignment loss to align the hash codes of each training sample to its associated hash center, supplementing global semantic guidance and enhancing sample efficiency.

We conduct extensive experiments on 4 datasets: ActivityNet, FCVID, UCF101, and HMDB51, demonstrating that \modelname{} outperforms state-of-the-art baselines under various setups and transfers better across datasets. 
Regarding inference efficiency, \modelname{} exhibits notable advantages, including lower memory overhead, which allows for larger batchsizes, and faster computation. 
As shown in \Cref{fig:intro}(a), these advantages become more pronounced as the frame number increases. 
Additionally, we provide comprehensive ablations and analyses, focusing on network architecture and training strategy. 
The results verify Mamba's superiority in SSVH and confirm the necessity of the global signal in the SLG learning paradigm. 
In particular, \Cref{fig:intro}(b) shows that the proposed center alignment loss, which serves as the global signal, can guide the training process toward faster and better convergence.

To sum up, our paper makes the following contributions.
\setlist{nolistsep}
\begin{itemize}[leftmargin=1.5em]
\item We explore the first Mamba-based SSVH model, indicating a superior solution for both efficacy and efficiency. 
\item We design a hash center generation algorithm that computes semantically consistent and discriminative hash centers from the feature-space global semantics.
\item We propose a center alignment loss as a global learning signal, contributing to a solid self-local-global (SLG) paradigm and improving training efficiency.
\end{itemize}

%% file: sections/RelatedWork.tex
\section{Related Works}
\label{sec:related_work}
\subsection{Self-Supervised Video Hashing} \label{subsec:ssvh_works}
Video hashing focuses on learning binary codes to enable fast, memory-efficient video retrieval. Self-supervised video hashing (SSVH) is particularly valuable for applications where labels are scarce. Early approaches often overlooked the temporal dynamics of videos, relying on traditional techniques like ITQ \cite{ITQ_2012}, SH \cite{SH_2008}, and MFH \cite{MFH_2011}. VHDT \cite{VHDT_2013} addressed this gap and showed notable improvement. 

Recent methods focused on deep models, with progress in network design and learning strategy. While RNNs prevailed \cite{SSTH_2016,JTAE_2017,SSVH_2018} initially, newer research \cite{BTH_2021,ConMH_2023,TSVH_2024} has favored Transformers \cite{vaswani2017attention} for superior performance. Other remarkable contributions include the use of MLP-Mixer \cite{MLPMixer_2021} in MCMSH \cite{MCMSH_2022} and EUVH \cite{EUVH_2024}, as well as the incorporation of graph networks \cite{GAT_2018} in MAGRH \cite{MAGRH_2022}.

On learning strategy, existing methods can be classified into 4 categories: 
(\textbf{i}) \emph{Self-recovery signals}, such as auto-regressive frame reconstruction \cite{SSTH_2016,SSVH_2018,UVVH_2019}, separated reconstructions of appearance and temporal dynamics \cite{JTAE_2017}, masked frame recovery \cite{BTH_2021,DKPH_2022,ConMH_2023}, and temporal order prediction \cite{CHAIN_2023}. 
(\textbf{ii}) \emph{Inter-sample \textbf{local} signals}, including pairwise similarity preservation \cite{tUSMVH_2017,SSVH_2018,NPH_2019,BTH_2021} and contrastive learning \cite{ConMH_2023,EUVH_2024}.
(\textbf{iii}) \emph{Regularization signals}, including hashing regularization techniques like minimizing quantization error, bit decorrelation, and bit balance \cite{wu2017unsupervised}, as well as novel methods such as self-distillation \cite{DKPH_2022}, temporal sensitivity regularization \cite{TSVH_2024}, bit-wise distribution prediction \cite{UVVH_2019,BerVAE_2023}, and feature-space cluster alignment as an auxiliary task \cite{BTH_2021,MCMSH_2022}.
(\textbf{iv}) \emph{Multi-modal signals}, where advanced methods incorporated extra modalities like motion \cite{MAGRH_2022,CTH_2023} or audio \cite{AVHash_2024}, benefiting from cross-modal alignment.

Our work presents two key contributions: 
(\textbf{i}) In network design, we are the first to explore the novel Mamba architecture \cite{Mamba_2023} in SSVH, achieving an optimal balance between performance and efficiency. 
(\textbf{ii}) In hash learning strategy, we introduce a \emph{global} signal through semantic hash center generation and a center alignment loss. 
Unlike previous methods \cite{BTH_2021,MCMSH_2022,EUVH_2024} that employed \emph{feature-space} cluster alignment as auxiliary regularization, our \emph{hash-space} center alignment is more direct and effective. 

\subsection{State Space Models} \label{subsec:ssm_works}
State space models (SSMs), originating from control theory \cite{kalman1960new}, have emerged as a powerful framework for sequence modeling. 
Recent research has focused on linear SSMs to improve efficiency, yielding representative works like HiPPO \cite{gu2020hippo} and LSSL \cite{LSSL_2021}. 
Based on this progress, the S4 model \cite{S4_2021} set a milestone with successful performance across various sequence tasks. 
Several following works further optimized S4 to balance efficacy and efficiency by replacing the diagonal plus low-rank structure with a simpler diagonal matrix \cite{S4D_2022}. 
Besides, many efforts have been devoted to hardware-efficient implementation. 
For example, the S5 model \cite{smith2022simplified} incorporated a MIMO implementation and efficient parallel scanning techniques. 
H3 \cite{fu2022hungry} addressed efficiency and performance gaps between SSMs and Transformers in language tasks, by proposing a fast FlashConv operator and a novel state passing algorithm. 

Among the advances in SSMs, Mamba \cite{Mamba_2023} made significant strides with a novel data-dependent selective mechanism and hardware-efficient implementation. 
The past few months have seen emerging interest in various Mamba-base applications, including but not limited to vision \cite{VMamba_2024, Vim_2024}, multi-modality \cite{VLMamba_2024,Cobra_2024}, and graph \cite{GraphMamba_2024}. 
Inspired by these successes, we take the first exploration of Mamba's potential in video hashing.

%% file: sections/Method.tex
\begin{figure*}[t]
    \centering
    \includegraphics[width=\textwidth]{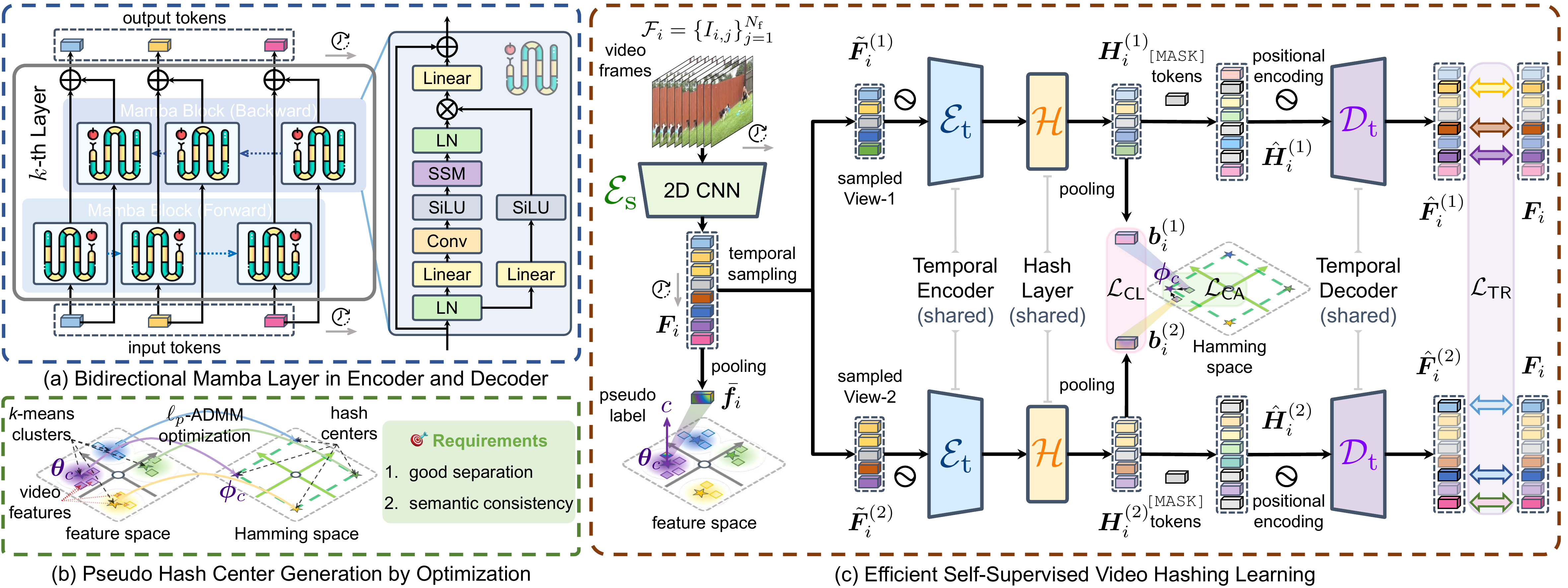}
    \caption{Overview of \modelname{} (best viewed in color).
    (\textbf{a}) The encoder and decoder comprise bidirectional Mamba layers for effective and efficient temporal modeling. 
    (\textbf{b}) We propose an optimization algorithm to transform the feature-space global structure into well-separated and semantically consistent hash centers. 
    (\textbf{c}) We encode video frames into features and get a pseudo label of the nearest feature cluster. 
    Then, we sample two views of the video and process them with the shared encoder and hash layer process, obtaining frame-wise soft hash vectors. 
    Next, we aggregate frame hash vectors to video-level hash vectors for contrastive learning and center alignment. 
    Meanwhile, we employ an auxiliary decoder (removed in inference) to reconstruct the masked frames, using the frame hash vectors of each view.}
    \label{fig:arc}
\end{figure*}

\section{Method}
\label{sec:method}

\subsection{Problem Formulation and Overview of \modelname{}}
\label{subsec:overview}
Suppose there is an unlabeled video corpus, $\calC=\dak{\bmF_i}_{i=1}^{N}$, where $\bmF_i\in\bbR^{N_t\times D}$ is the frame feature collection of the $i$-th video, extracted by pre-trained 2D CNNs \cite{simonyan2014very,he2016deep}. $N_t$ and $D$ denote frame number and feature dimension, respectively. 
Self-supervised video hashing (SSVH) aims to take $\bmF_i$ as input and generate a hash vector $\bmb_i\in\{-1,+1\}^K$, such that the Hamming distance can precisely reflect the semantic similarity. 
For this task, we propose \textbf{\modelname{}}, a Mamba-based network trained with an improved paradigm, as illustrated in \Cref{fig:arc}.

\subsection{Mamba-based Video Hash Network} \label{subsec:mamba_layer}
\subsubsection{Preliminaries: State-Space Models (SSMs) and Mamba} \label{subsec:mamba_basic}
SSMs map input $x(t)\in\mathbb{R}^L$ to output $y(t)\in\mathbb{R}^L$ via the hidden state $h(t)\in\mathbb{R}^N$. Here, $\bmA \in \mathbb{R}^{N \times N}$ defines the hidden state's evolution, while $\bmB \in \mathbb{R}^{N \times 1}$ and $\bmC \in \mathbb{R}^{1 \times N}$ represent the input and output mappings, respectively. We can express it by linear ordinary differential equations (ODEs):
\begin{equation}
\begin{aligned} \label{equ:ODE}
\bmh'(t)&=\bmA \bmh(t)+\bmB \bmx(t), \\ 
\bmy(t)&=\bmC \bmh(t).
\end{aligned}
\end{equation}

Modern SSMs approximate ODEs by discretizing $\bmA$ and $\bmB$ using a timescale $\Delta$, through zero-order hold:
\begin{gather}
\bar{\bmA}=\exp(\Delta\bmA), \\
\bar{\bmB}=(\Delta\bmA)^{-1}(\exp(\Delta\bmA)-\bmI)\cdot\Delta\bmB,
\end{gather}
so that we obtain the discretized version of \Cref{equ:ODE}:
\begin{equation}
\begin{aligned}
\bmh_t&=\bar{\bmA}\bmh_{t-1}+\bar{\bmB}\bmx_t, \\
\bmy_t&=\bmC \bmh_t.
\end{aligned}
\end{equation}

Mamba~\cite{Mamba_2023} introduced data dependence to $\Delta$, $\bmB$, and $\bmC$, enabling input-aware selection for improved modeling. Despite limitations in parallelism due to recurrence, Mamba enhanced efficiency through structural reparameterization and parallel scanning algorithms.

\subsubsection{Bidirectional Mamba Layers} \label{subsubsec:bd_mamba_layer}
To effectively extract semantic information in videos, we build the temporal encoder $\calE_t$ and decoder $\calD_t$ with multi-layer bidirectional Mamba \cite{Mamba_2023} layers. 
As illustrated in \Cref{fig:arc}(a), each layer is composed of a \emph{forward} Mamba block and a \emph{backward} (\ie \emph{reverse}) Mamba block, which is formulated by 
\begin{equation}
    \bmS_\text{out} = \overrightarrow{\mathrm{Mamba}}(\bmS_\text{in}) + \overleftarrow{\mathrm{Mamba}}(\bmS_\text{in}). 
\end{equation}
Here, $\rightarrow$ and $\leftarrow$ mark the forward and the reverse scans, respectively. 
$\bmS_\text{in}$ and $\bmS_\text{out}$ denote the input and the output hidden states, respectively. 
Each Mamba block adopts a gated structure with two branches, for example, the forward block:
\begin{gather}
    \overrightarrow{\bmS}_\mathrm{out} = \mathrm{Linear}_3(\overrightarrow{\bmS}' \otimes \bmS''), \\
    \overrightarrow{\bmS}' = \mathrm{LN}_2(\overrightarrow{\mathrm{SSM}}(\sigma(\mathrm{Conv}(\mathrm{Linear}_1(\mathrm{LN}_1(\bmS_\mathrm{in})))))), \\
    \bmS'' = \sigma(\mathrm{Linear}_2(\bmS_\mathrm{in})),
\end{gather}
where $\overrightarrow{\bmS}'$ and $\bmS''$ are hidden states of the main and the gating branches, respectively. 
$\otimes$ denotes the Hadamard product. 
$\mathrm{LN}$ denotes layer normalization. 
$\sigma$ denotes the SiLU activation \cite{SiLU_2017}.
$\overrightarrow{\mathrm{SSM}}$ denotes the forward selective scan module.
$\mathrm{Conv}$ denotes the 1D convolution. 
$\mathrm{Linear}$ denotes learnable linear projection. 

\subsubsection{Hash Layer} \label{subsubsec:hash_layer}
We design a hash layer upon encoder's output to transform visual embeddings into compact hash vectors. 
Given the encoded embeddings of the $i$-th video, $\bmE_i\in\bbR^{N_t\times D}$, we obtain $K$-dimensional soft hash vectors by 
\begin{gather}
    \label{equ:frame_hash_codes}
    \bmH_i = \mathrm{tanh}(\mathrm{Linear}(\bmE_i))\in(-1,+1)^{N_t\times K},
\end{gather}
where $\mathrm{tanh}$ denotes the hyperbolic tangent function. To establish video-level hash codes, we aggregate the frame hash codes by mean pooling and the $\mathrm{sign}$ function, namely
\begin{equation}
    \label{equ:video_hash_codes}
    \bmb_i = \mathrm{sign}\biggl(\frac{1}{N_t}\sum_{j=1}^{N_t}\bmH_i[j]\biggr)\in\{-1,+1\}^K. 
\end{equation}
For end-to-end training, we pass the gradient through \cite{STE_2013} the $\mathrm{sign}$ function.

\subsection{Semantic Hash Center Generation} \label{subsubsec:center_generation}
Existing SSVH methods found it hard to use feature-space semantics as effectively as in supervised scenarios. 
They can only take inter-sample similarity as proxy or regularize feature space to assist hash learning, showing limited training efficiency. 
We break this dilemma with a more direct solution, exploring global categorical semantics for improved hash learning. 
For this purpose, we introduce how to (\textbf{i}) \emph{extract} and (\textbf{ii}) \emph{transform} such implicit information into the hash space. 
This process can be done before model training.

\subsubsection{Global Semantic Structure Extraction} \label{subsubsec:global_semantic}
Considering the large-scale clustering on video corpus, rather than all frame features, we use the temporally averaged features of videos to reduce the order of magnitude. 
By the $k$-means algorithm, we obtain $N_c$ cluster centroids, $\bmTheta = [\opseq{\bmtheta}{N_c}{;}]$, $\bmtheta_*\in\bbR^{D}$, which can be regarded as the compression of the corpus and encode the global semantic structure. 

\subsubsection{Optimization for Hash Center Generation} \label{subsec:hash_centers}
We further transform $\bmTheta$ into hash centers, which are expected to be 
(\textbf{i}) semantically consistent with the feature space, and 
(\textbf{ii}) well separated from each other to encourage discriminative hash codes. 
Let us translate these requirements into objectives:
\begin{gather}
    \notag
    \min_{\bmPhi} \overbrace{\shuk{\bmPhi\bmPhi^\top - K\bmW}_\mathrm{F}^2}^\text{semantic consistency} + \overbrace{\frac{1}2\sum_{i,j}^{N_c}\bmphi_i^\top\bmphi_j}^\text{separation}, \\
    \label{equ:center_optim_problem}
    \textrm{s.t.}\ \bmPhi = [\opseq{\bmphi}{N_c}{;}]\in\{-1,+1\}^{N_c\times K}.
\end{gather}
$\bmW_{ij} = \cos(\bmtheta_i,\bmtheta_j)$ is the feature-space semantic similarity. 
$\bmPhi$ denotes the desired hash center collection. 

It is NP-hard to optimize \Cref{equ:center_optim_problem} for the binary constraint of hash centers. Fortunately, we take inspiration from \citet{LpBoxADMM_2018} that $\bmPhi\in\{-1,+1\}^{N_c\times K}$ is equivalent to $\bmPhi\in[-1,+1]^{N_c\times K}\cap\{\bmPhi\mid\shuk{\bmPhi}_p^p=N_cK\}$ and adopt the $\ell_p$-box ADMM algorithm to solve \Cref{equ:center_optim_problem}. 

Concretely, we first introduce two auxiliary variables $\bmPsi_b$ and $\bmPsi_p$ associated with the constrains $\calS_b\equiv[-1,+1]^{N_c\times K}$ and $\calS_p\equiv\{\bmPsi_p\mid\shuk{\bmPsi_p}_p^p=N_cK\}$, respectively. 
Then, we solve the following problem with $p=2$ for simplicity: 
\begin{equation}
\label{equ:admm_problem}
\begin{aligned}
    \min_{\bmPhi,\bmPsi_*,\bmUpsilon_*} &\shuk{\bmPhi\bmPhi^\top - K\bmW}_\mathrm{F}^2 + \frac{1}2\sum_{i,j}^{N_c}\bmphi_i^\top\bmphi_j \\ 
    &+ \delta_{\calS_b}(\bmPsi_b) + \delta_{\calS_p}(\bmPsi_p) \\
    &+\frac{\mu_b}2\shuk{\bmPhi-\bmPsi_b}^2_\mathrm{F}+\frac{\mu_p}2\shuk{\bmPhi-\bmPsi_p}^2_\mathrm{F} \\
    &+\mathrm{Tr}(\bmUpsilon_b^\top(\bmPhi-\bmPsi_b))+\mathrm{Tr}(\bmUpsilon_p^\top(\bmPhi-\bmPsi_p)).
\end{aligned}
\end{equation}
$\delta_\calS(\bmPsi)$ is an indicator that outputs $0$ if $\bmPsi\in\calS$ else $+\infty$. $\bmUpsilon_*$ and $\mu_*$ are the dual and penalty variables, respectively. 
Next, the optimization process follows \citet{LpBoxADMM_2018}. 

\textbf{Update $\bmPhi$:} We fix all variables except for $\bmPhi$ at the $(k+1)$-th iteration. $\bmPhi^{k+1}$ is updated by
\begin{equation}
\resizebox{.87\hsize}{!}{
    $\min_{\bmPhi} \shuk{\bmPhi\bmPhi^\top - K\bmW}_\mathrm{F}^2 + \frac{1}2\sum_{i,j}^{N_c}\bmphi_i^\top\bmphi_j + \frac{\mu_b+\mu_p}2\shuk{\bmPhi}_\mathrm{F}^2+\mathrm{Tr}(\bmPhi\bmG^\top)$,
}
\end{equation}
where $\bmG = \bmUpsilon_b^k+\bmUpsilon_p^k - \mu_b\bmPsi_b^k - \mu_p\bmPsi_p^k$. 
The gradient can be calculated by the LBFGS-B method as
\begin{equation}
    4(\bmPhi\bmPhi^\top - K\bmW)\bmPhi + \bmone\bmone^\top\bmPhi + (\mu_b+\mu_p)\bmPhi+\bmG.
\end{equation}

\textbf{Update $\bmPsi_*$:} We fix $\bmPhi^{k+1}$ and $\bmUpsilon_b^k$, updating $\bmPsi_b^{k+1}$ by 
\begin{equation}
    \min_{\bmPsi_b} \delta_{\calS_b}(\bmPsi_b) + \frac{\mu_b}{2}\shuk{\bmPsi_b-\bmPhi^{k+1}}_\mathrm{F}^2-\mathrm{Tr}({\bmUpsilon_b^k}^\top\bmPsi_b),
\end{equation}
which can be easily solved with the proximal minimization method, yielding the closed-form solution: 
\begin{equation}
    \bmPsi^{k+1}_b = \sqrt{N_cK}\times\frac{\bmPhi^{k+1}+\bmUpsilon_b^k/\mu_b}{\shuk{\bmPhi^{k+1}+\bmUpsilon_b^k/\mu_b}_\mathrm{F}}.
\end{equation}
Updating $\bmPsi_p^{k+1}$ follows the analogous procedure.

\textbf{Update $\bmUpsilon_*$:} We update them by gradient ascent:
\begin{gather} 
    \label{equ:update_bmUpsilon_b}
    \bmUpsilon_b^{k+1} = \bmUpsilon_b^k + \eta\mu_b (\bmPhi^{k+1}-\bmPsi_b^{k+1}), \\
    \label{equ:update_bmUpsilon_p}
    \bmUpsilon_p^{k+1} = \bmUpsilon_p^k + \eta\mu_p (\bmPhi^{k+1}-\bmPsi_p^{k+1}),
\end{gather}
where $\eta$ is the learning rate. 
The above optimization process alternates between each variable until convergence. 

\subsection{Self-Local-Global (SLG) Learning Paradigm} \label{subsec:SLG_paradigm}
To enhance hash learning, we faithfully leverage hierarchical learning signals in different considerations, including 
(\textbf{i}) temporal reconstruction as \emph{self-recovery} signal to capture relations in temporal dynamics, 
(\textbf{ii}) contrastive learning as \emph{inter-sample local} signal for discriminative hash codes, 
and (\textbf{iii}) hash center alignment as \emph{inter-sample global} signal to prompt faster, better convergence. 
They each play indispensable roles in efficient and effective learning.

\subsubsection{Temporal Reconstruction} 
\label{subsubsec:temporal_recon}
Following \citet{ConMH_2023}, we reconstruct the masked frame features from the frame-level hash codes to maximize their semantic capacity. 
Specifically, we take the frame hash codes of the augmented view $n$ of the $i$-th video, namely, $\bmH^{(n)}_i$, as a showcase. 
It has dropped the frames associated with indices $\calM^{(n)}_i$ during data augmentation. 
We first insert the \texttt{[mask]} token (a learnable vector in $\bbR^K$) to $\bmH^{(n)}_i$ according to $\calM^{(n)}_i$ and obtain the decoder input, $\hat{\bmH}^{(n)}_i$. 
Then we process it with the temporal decoder $\calD_t$ and get decoded features by $\hat{\bmF}_i^{(n)} = \calD_t(\hat{\bmH}_i^{(n)})$. 
Finally, we compute reconstruction loss for the masked frames as
\begin{gather}
    \calL_\mathsf{TR}^{(n)} = \frac{1}{|\calM_i^{(n)}|}\sum_{m\in\calM_i^{(n)}}\|\hat{\bmF}_i^{(n)}[m] - \bmF_i[m]\|^2_2. 
\end{gather}

\subsubsection{Contrastive Learning} \label{subsubsec:contrastive_learn}
We contrastively align video-level hash codes between views, with a temperature factor $\tau>0$:
\begin{gather}
\label{equ:contrastive_loss}
\resizebox{.87\hsize}{!}{
    $\calL_\mathsf{CL} = -\log\frac{\exp(\cos(\bmb_i^{(1)}, \bmb_i^{(2)})/\tau)}{\sum_{j}\exp(\cos(\bmb_i^{(1)}, \bmb_j^{(2)})/\tau)}\cdot\frac{\exp(\cos(\bmb_i^{(1)}, \bmb_i^{(2)})/\tau)}{\sum_{j}\exp(\cos(\bmb_j^{(1)}, \bmb_i^{(2)})/\tau)}$.
}
\end{gather}

\subsubsection{Hash Center Alignment} \label{subsubsec:center_alignment}
We obtain the pseudo label $c_i$ for the $i$-th video by clustering its temporally averaged features, then align the video hash codes $\bmb_i^{(n)}$ to the hash center of $c_i$:
\begin{equation}
\label{equ:center_loss}
\begin{aligned}
    \calL_\mathsf{CA}^{(n)} &= -\log\frac{\exp(\cos(\bmphi_{c_i},\bmb_i^{(n)})/\tau)}{\sum_{c=1}^{N_c}\exp(\cos(\bmphi_{c},\bmb_i^{(n)})/\tau)} \\
    &= -\log\frac{\exp(\bmphi_{c_i}^\top\bmb_i^{(n)}/K\tau)}{\sum_{c=1}^{N_c}\exp(\bmphi_{c}^\top\bmb_i^{(n)}/K\tau)}.
\end{aligned}
\end{equation}

\subsubsection{Total Learning Objectives} 
\label{subsubsec:total_loss}
\begin{equation}
    \calL_\text{S5VH} = \frac12(\calL_\mathsf{TR}^{(1)}+\calL_\mathsf{TR}^{(2)}) + \alpha\calL_\mathsf{CL} + \frac\beta2(\calL_\mathsf{CA}^{(1)}+\calL_\mathsf{CA}^{(2)}). 
\end{equation}
$\alpha, \beta>0$ are hyperparameters to balance learning signals. 

%% file: sections/Experiments.tex
\section{Experiments}
\label{sec:experiments}
\subsection{Experimental Setup}
\subsubsection{Datasets}
We conduct experiments on 4 benchmark datasets.
(\textbf{i}) \textbf{ActivityNet} \cite{caba2015activitynet} contains 200 activity categories of recognition. 
We follow the standard setup as in \citet{ConMH_2023}, using 9,722 videos for training.
We uniformly sample 1,000 videos across 200 categories in the validation set as queries, and the remaining 3,758 videos as the database. 
(\textbf{ii}) \textbf{FCVID} \cite{jiang2017exploiting} contains 91,223 videos across 239 categories. 
We follow \citet{song2018self} to use 45,585 videos for training and 45,600 videos for the retrieval database and queries. 
(\textbf{iii}) \textbf{UCF101} \cite{UCF-1012012arxiv} consists of 13,320 videos from 101 human actions. 
We use 9,537 videos for training and the database, and 3,783 videos from the test set as the query set. 
(\textbf{iv}) \textbf{HMDB51} \cite{HMDB2011ICCV} comprises 6,849 videos across 51 actions. We use 3,570 videos for both training and database and 1,530 videos from the test set are designated as the query set.

\subsubsection{Metrics}
Following previous works \cite{MCMSH_2022,ConMH_2023}, we use mean Average Precision at top-$N$ results (\textbf{mAP@$N$}) as the metric, namely
\begin{equation}
    \textbf{mAP@}N = \frac1{Q}\sum_{q=1}^{Q}\text{AP@}N(q).
\end{equation}
Here, $Q$ is the number of queries in evaluation. AP@$N(q)$ means the Average Precision for the $q$-th query, defined by
\begin{equation}
    \text{AP}@N=\frac1{|\text{Rel}(N)|}\sum^N_{n=1}P(n)\cdot r(n), 
\end{equation}
where $|\text{Rel}(N)|$ is the total amount of relevant items. 
$P(n)$ denotes the precision at the $n$-th position. 
$r(n)$ is the relevance of the $n$-th ranked item ($0$: irrelevant; $1$: relevant).
We set $N\in\{5,20,40,60,80,100\}$ as showcases.
We further compute the geometric mean of these showcases, denoted as \textbf{GmAP} for simplicity, to show the holistic performance:
\begin{equation}
    \textbf{GmAP}=\sqrt{\sum_{N\in\{5,20,40,60,80,100\}}(\text{mAP@}N)^2}.
\end{equation}

Additionally, we use \textbf{Precision-Recall (PR)} curves to illustrate the detailed performance.

\subsection{Implementation Details}
\subsubsection{Frame Encoding}
For ActivityNet, we sample 30 frames per video and use ResNet-50 \cite{he2016deep} to extract 2048-D features. 
Both CNNs are pre-trained on ImageNet. 
For UCF101, HMDB51, and FCVID, we uniformly sample 25 frames per video and use VGG-16 \cite{simonyan2014very} to extract 4096-D features. 

\subsubsection{Model Configurations}
Considering the bidirectional layers yield double cost, to keep a comparable model size to baselines \cite{ConMH_2023}, we set the $6$ layers for the encoder and $1$ layer for the decoder. 
The latent dimensions of the encoder and decoder are set to $256$ and $192$, respectively. 

\subsubsection{Training Configurations}
For the model training, we choose the AdamW optimizer with default parameters in Pytorch, and employ a cosine annealed learning rate scheduling from $5e^{-4}$ to $1e^{-5}$. 
The models are trained for up to 350 epochs with 5-patience early-stopping to prevent over-fitting. 
The default hyperparameter configurations are as below: 
(\textbf{i}) We set the mask ratio $\rho=|\calM|/N_t$ to $0.75$ on the FCVID dataset and $0.5$ on the rest of the datasets. 
(\textbf{ii}) The temperature factor $\tau$ in \Cref{equ:contrastive_loss,equ:center_loss} is set $0.5$. 
(\textbf{iii}) The number of semantic centers $N_c$ is set to $450$ on FCVID and $100$ on the other datasets. 

\begin{figure}[t]
\centering
\subfloat{\includegraphics[width = 0.48\textwidth]{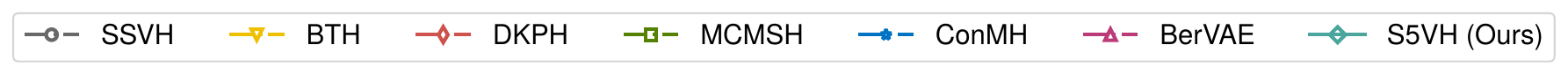}}
\setcounter{subfigure}{0}
\subfloat[\small{ActivityNet 16 bits}]{\includegraphics[width = 0.1600\textwidth]{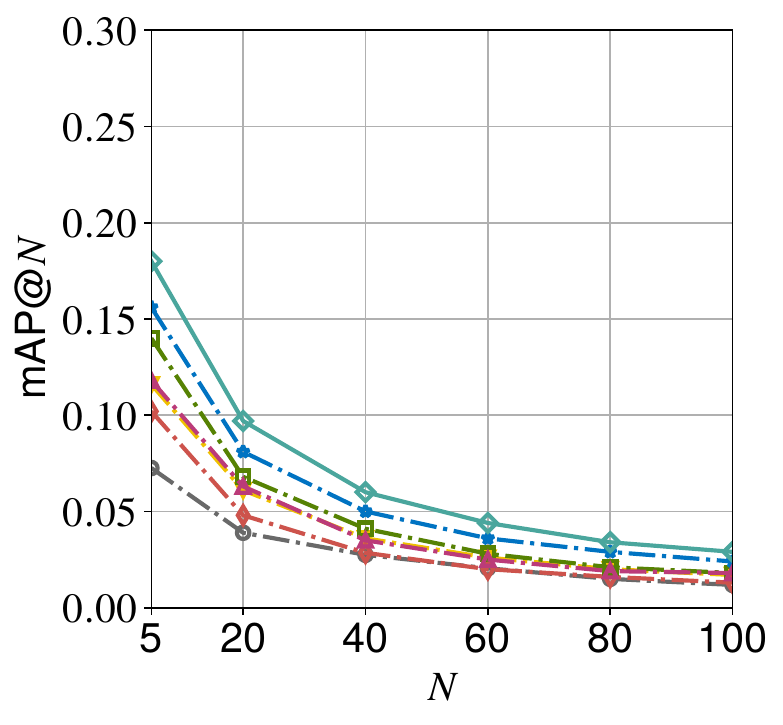}}
\subfloat[\small{ActivityNet 32 bits}]{\includegraphics[width = 0.1600\textwidth]{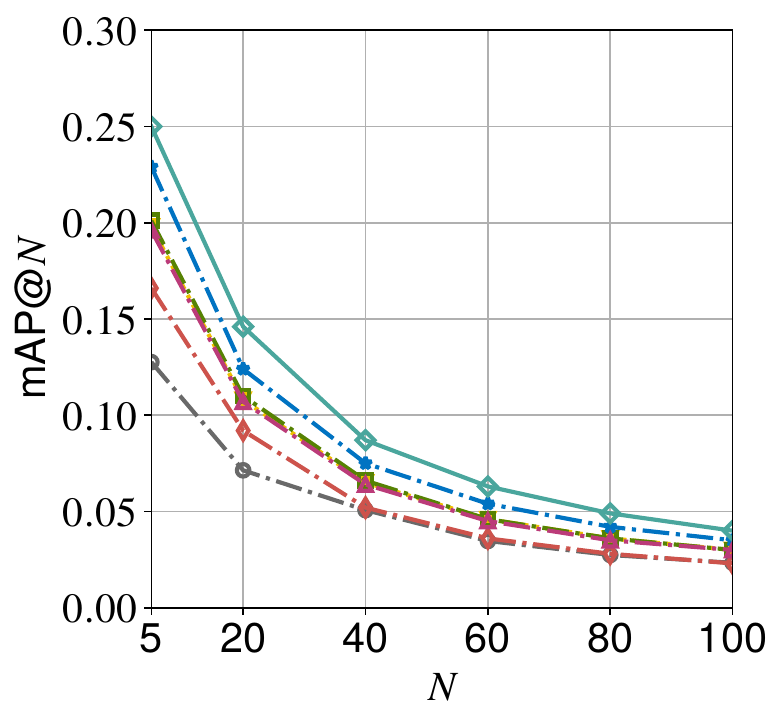}}
\subfloat[\small{ActivityNet 64 bits}]{\includegraphics[width = 0.1600\textwidth]{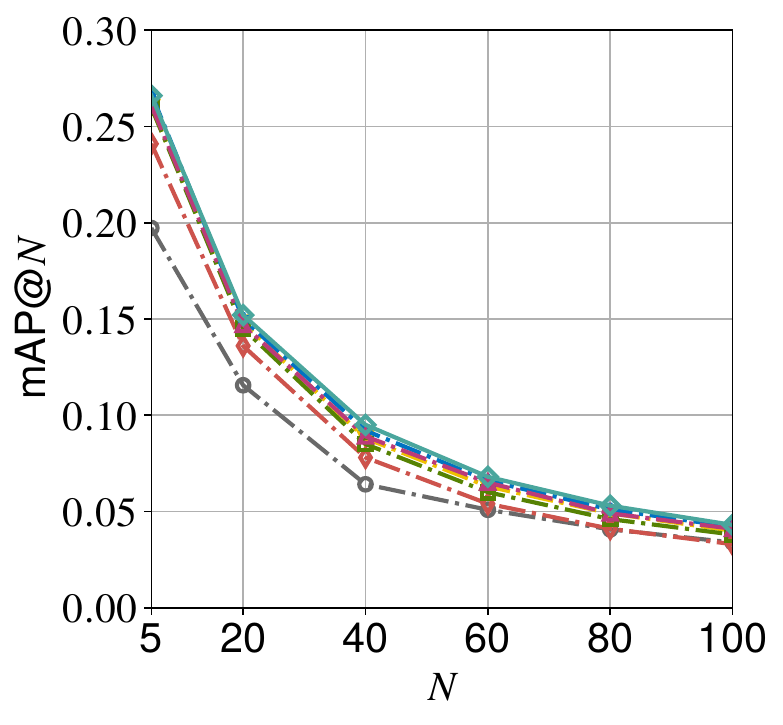}}

\subfloat[\small{FCVID 16 bits}]{\includegraphics[width = 0.1600\textwidth]{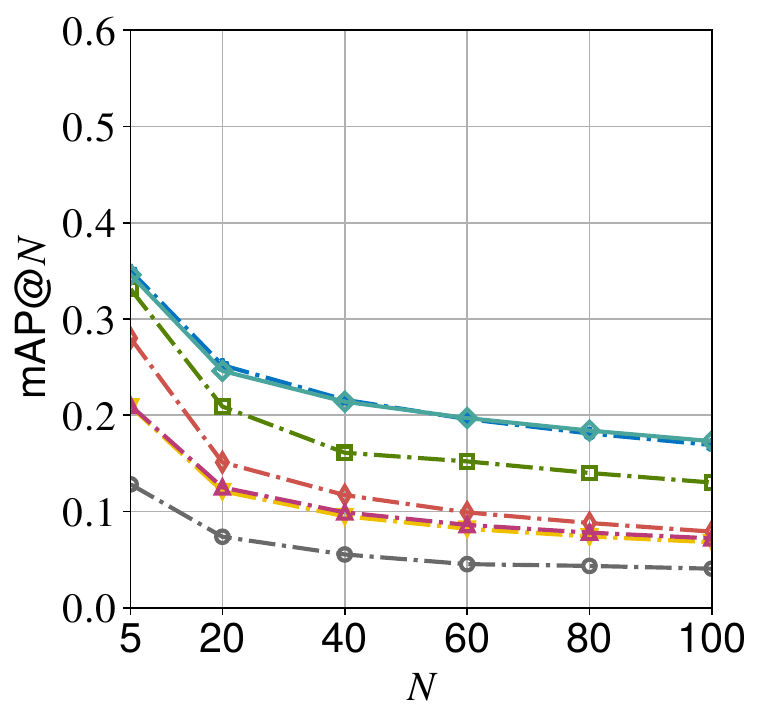}}
\subfloat[\small{FCVID 32 bits}]{\includegraphics[width = 0.1600\textwidth]{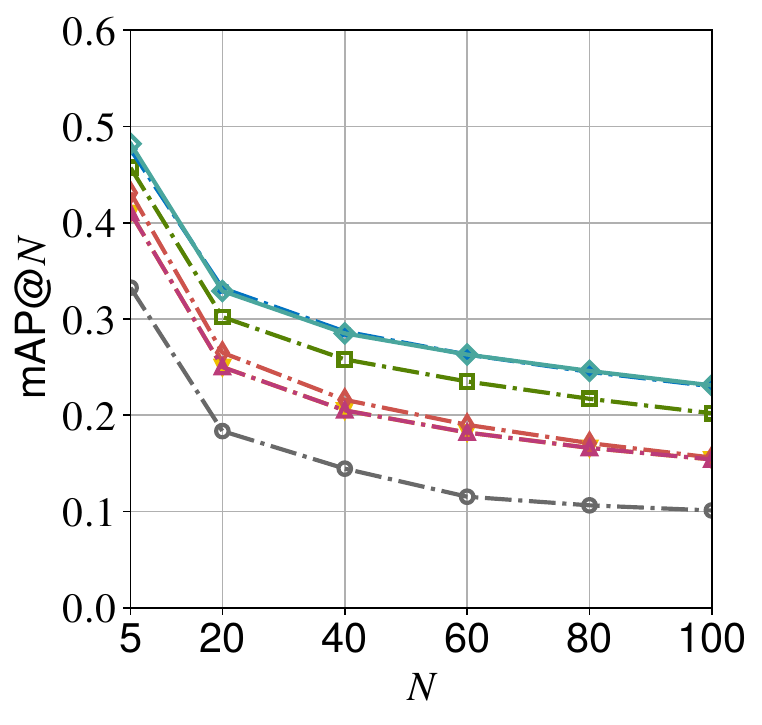}}
\subfloat[\small{FCVID 64 bits}]{\includegraphics[width = 0.1600\textwidth]{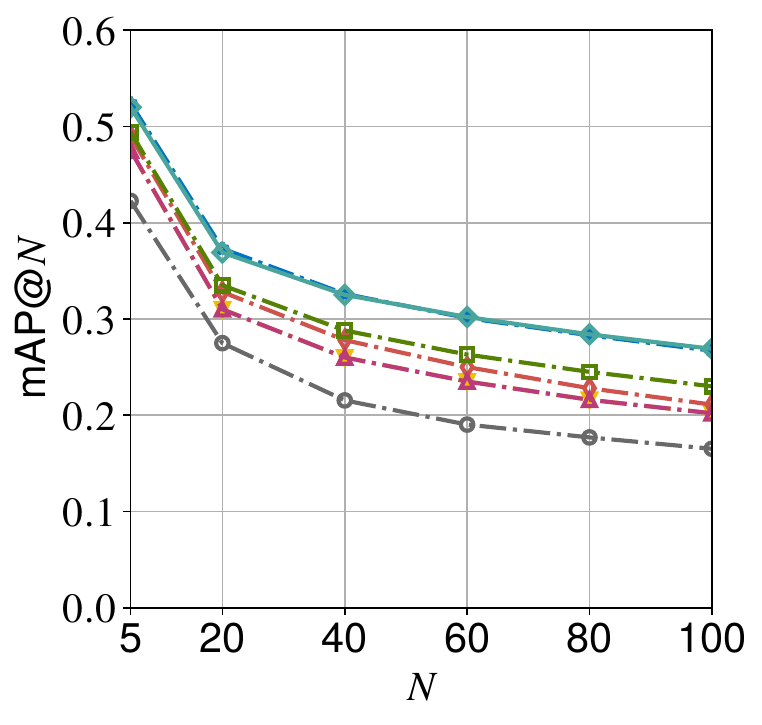}}

\subfloat[\small{UCF101 16 bits}]{\includegraphics[width = 0.1600\textwidth]{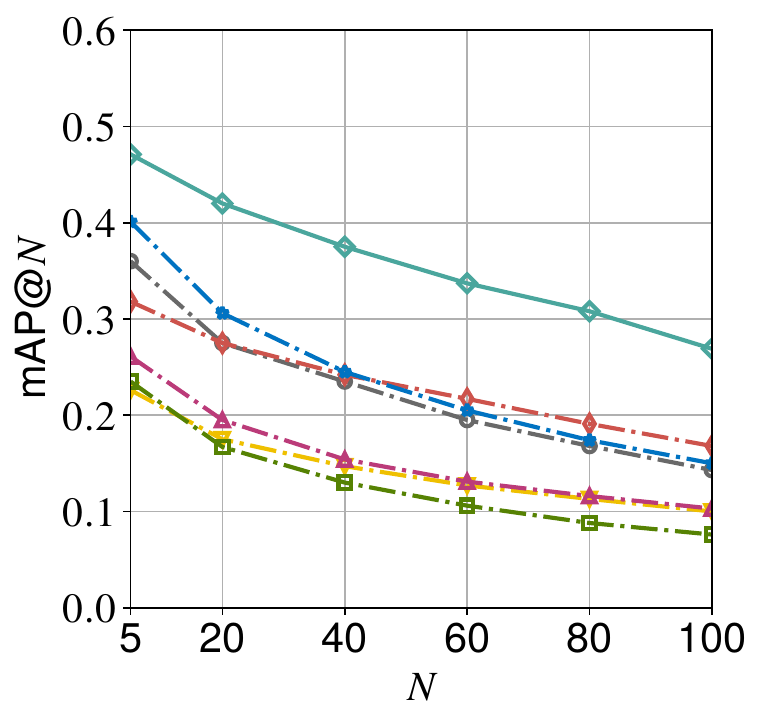}}
\subfloat[\small{UCF101 32 bits}]{\includegraphics[width = 0.1600\textwidth]{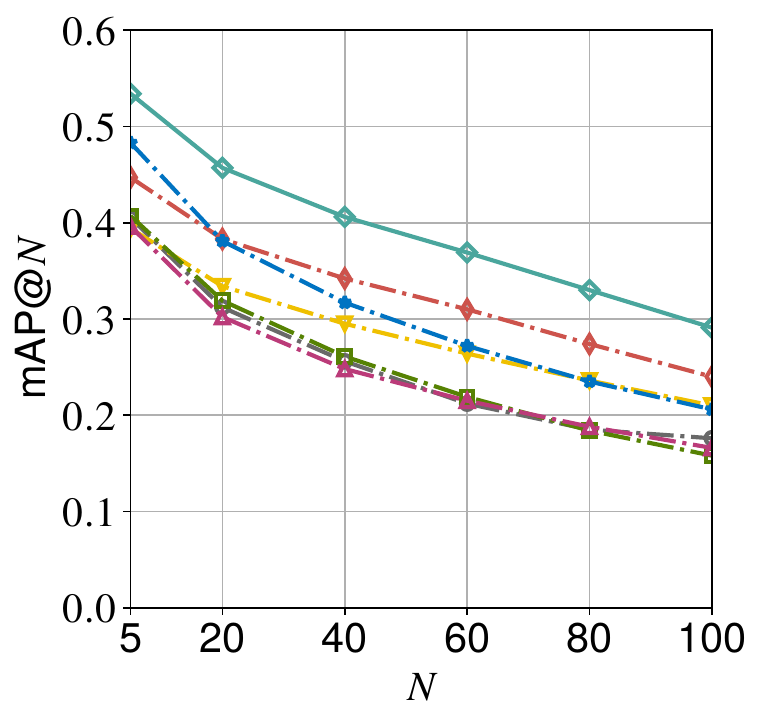}}
\subfloat[\small{UCF101 64 bits}]{\includegraphics[width = 0.1600\textwidth]{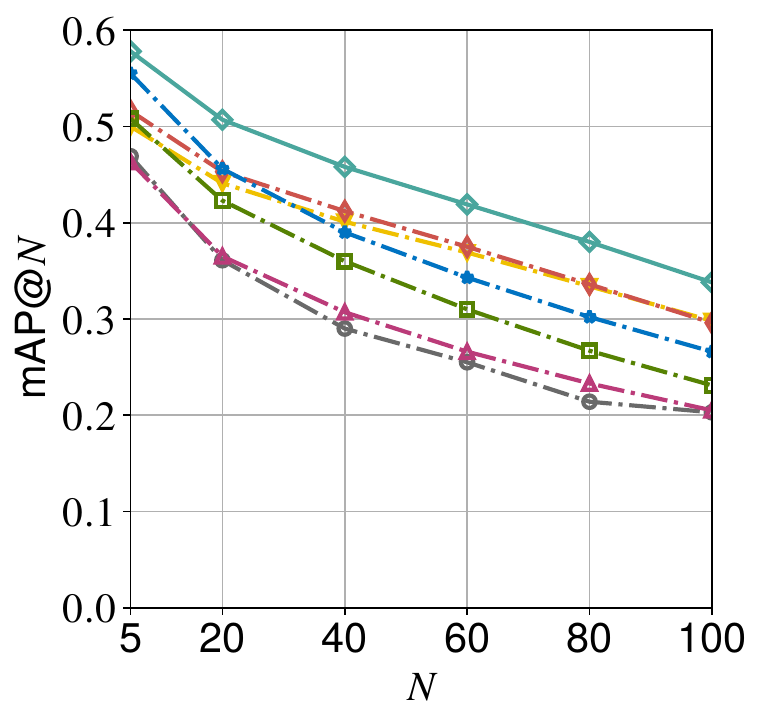}}

\subfloat[\small{HMDB51 16 bits}]{\includegraphics[width = 0.1600\textwidth]{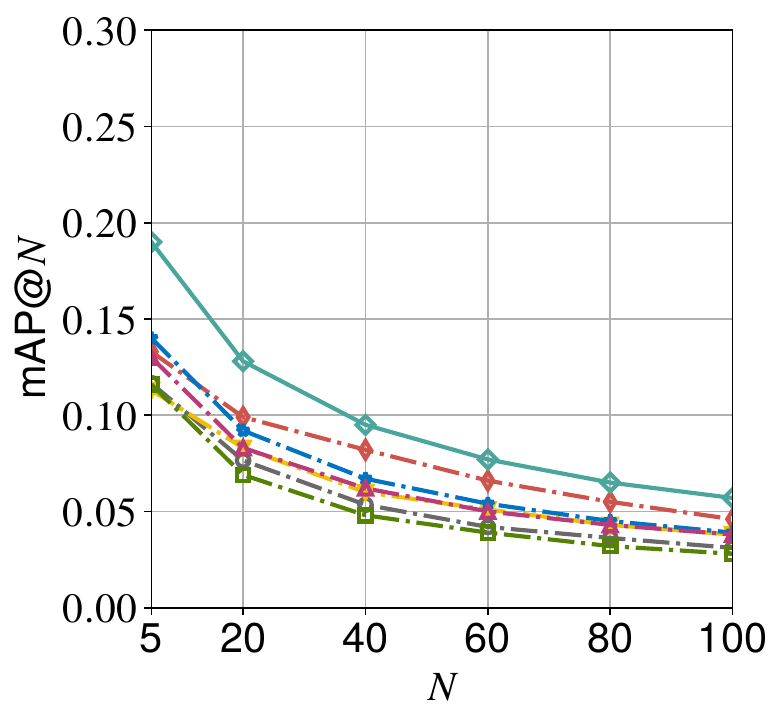}}
\subfloat[\small{HMDB51 32 bits}]{\includegraphics[width = 0.1600\textwidth]{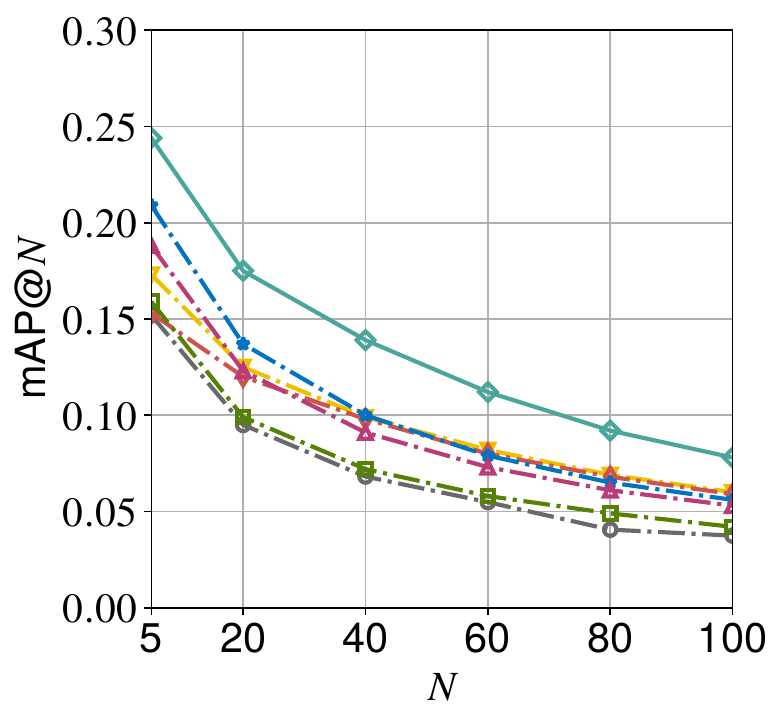}}
\subfloat[\small{HMDB51 64 bits}]{\includegraphics[width = 0.1600\textwidth]{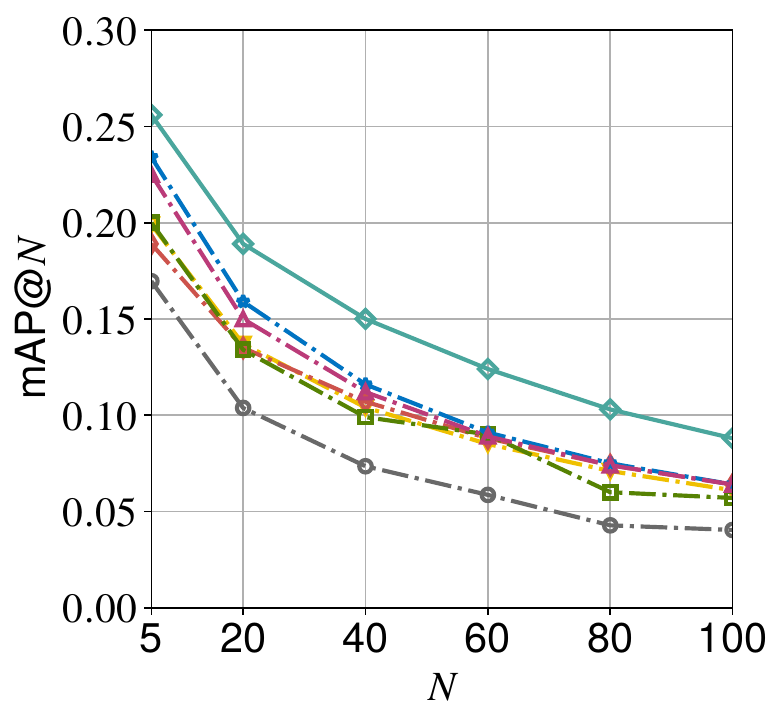}}
\caption{Retrieval performance comparison by mAP@$N$.}
\label{fig:MAP_curve}
\end{figure}

\subsection{Comparison with State-of-the-arts}
\subsubsection{Baselines}
We select 6 representative baselines for comparison:
SSVH \cite{SSVH_2018}, 
BTH \cite{BTH_2021}, 
DKPH \cite{DKPH_2022}, 
MCMSH \cite{MCMSH_2022}, 
ConMH \cite{ConMH_2023} and 
BerVAE \cite{BerVAE_2023}. 
We have discussed them in the Related Works section.

\subsubsection{Performance under Standard Protocols}
As illustrated in \Cref{fig:MAP_curve}, \modelname{} generally outperforms other methods across datasets and code lengths, demonstrating a superior efficacy. 
In particular, the improvements are more pronounced with lower-bit settings such as 16 bits, highlighting \modelname{}'s advantages in scenarios where high top-ranked results and retrieval speed are crucial.

\begin{figure}[t]
\centering
\subfloat{\includegraphics[width = 0.48\textwidth]{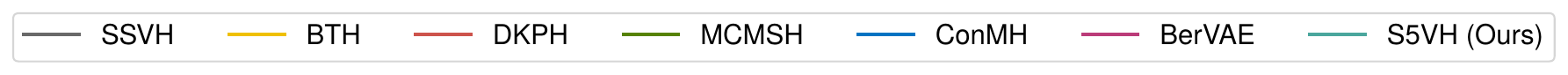}}
\setcounter{subfigure}{0}
\subfloat[\small{UCF101 16 bits}]{\includegraphics[width = 0.1600\textwidth]{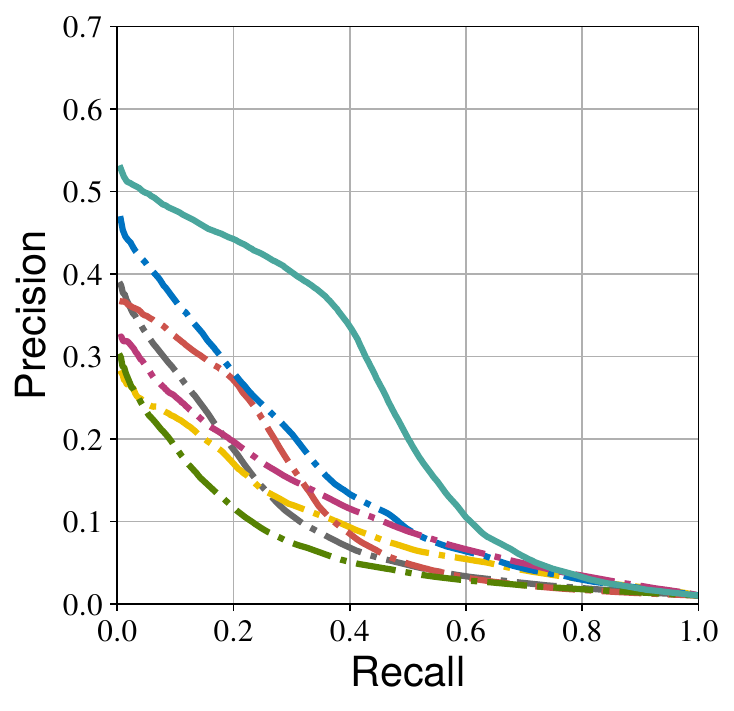}}
\subfloat[\small{UCF101 32 bits}]{\includegraphics[width = 0.1600\textwidth]{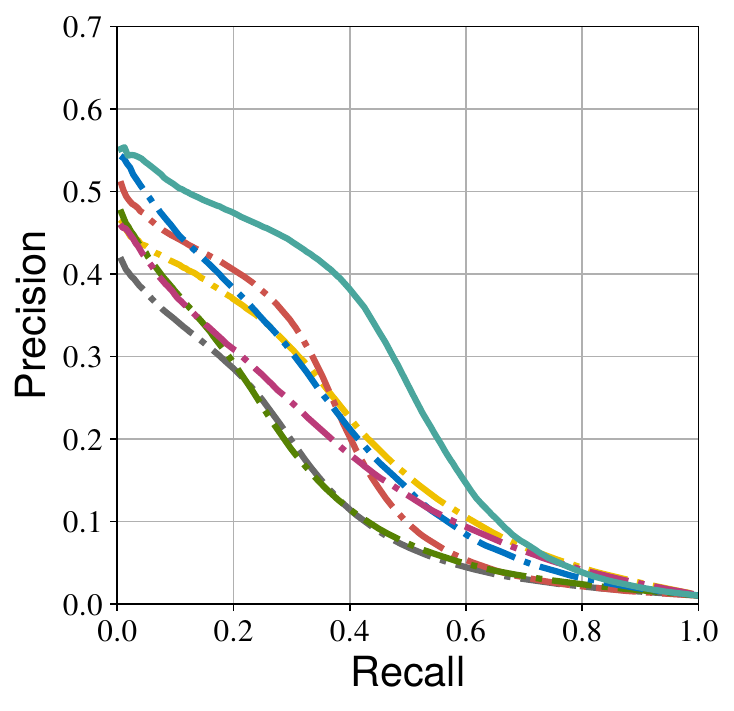}}
\subfloat[\small{UCF101 64 bits}]{\includegraphics[width = 0.1600\textwidth]{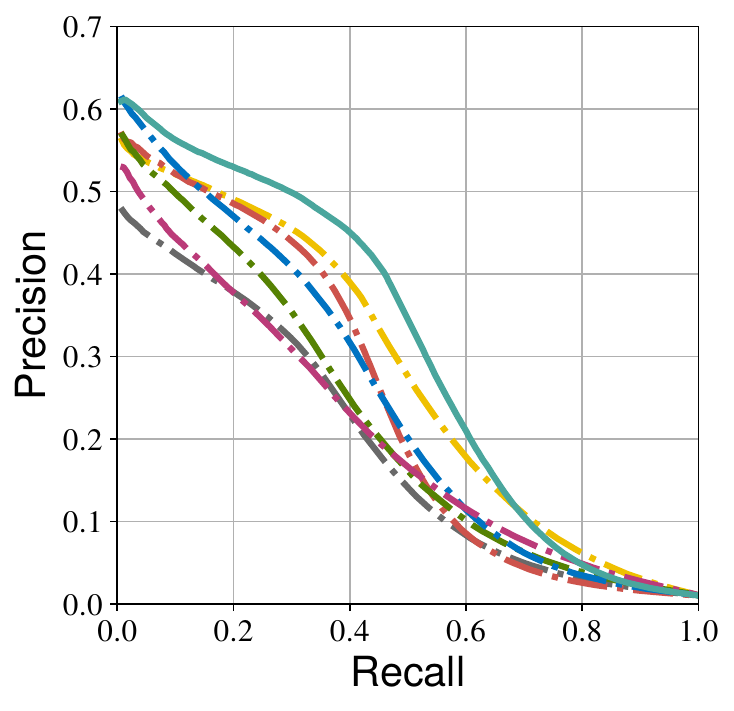}}

\subfloat[\small{HMDB51 16 bits}]{\includegraphics[width = 0.1600\textwidth]{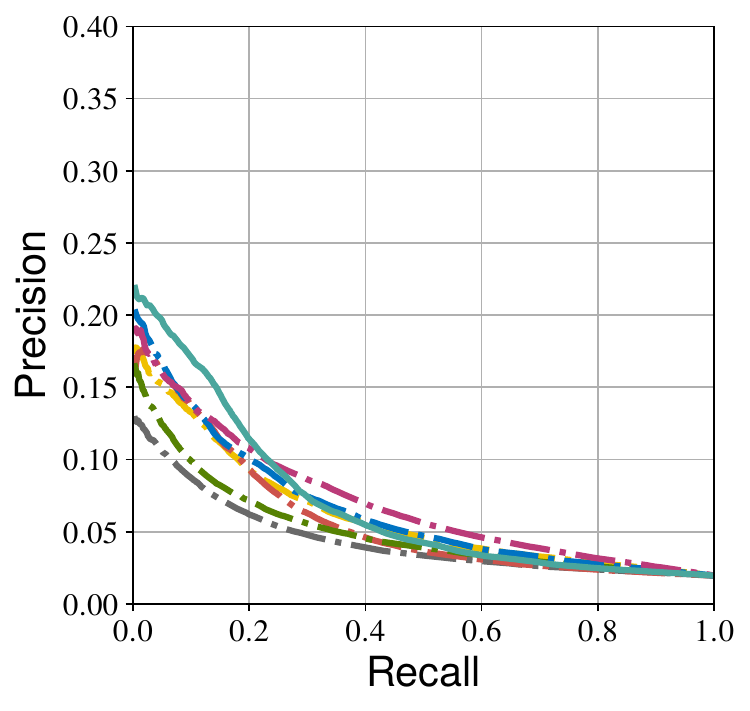}}
\subfloat[\small{HMDB51 32 bits}]{\includegraphics[width = 0.1600\textwidth]{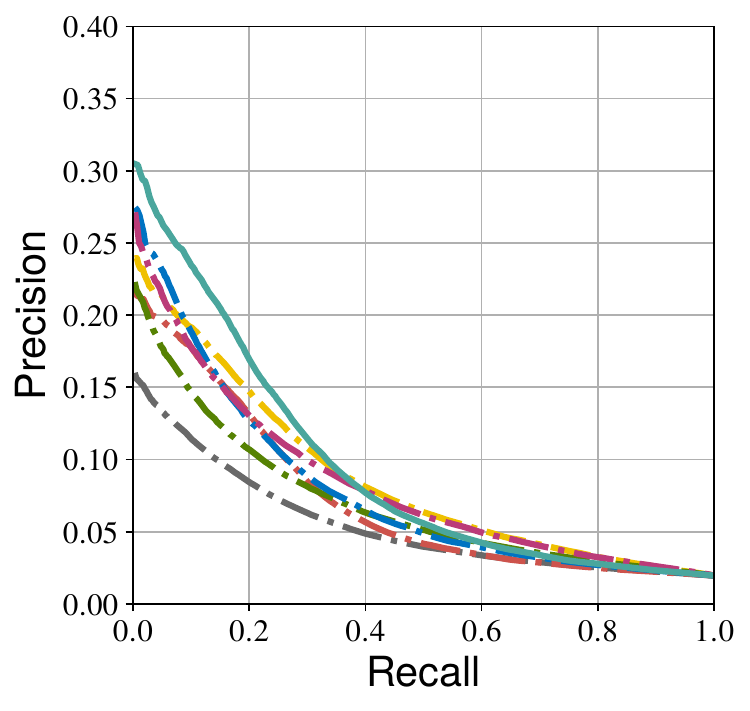}}
\subfloat[\small{HMDB51 64 bits}]{\includegraphics[width = 0.1600\textwidth]{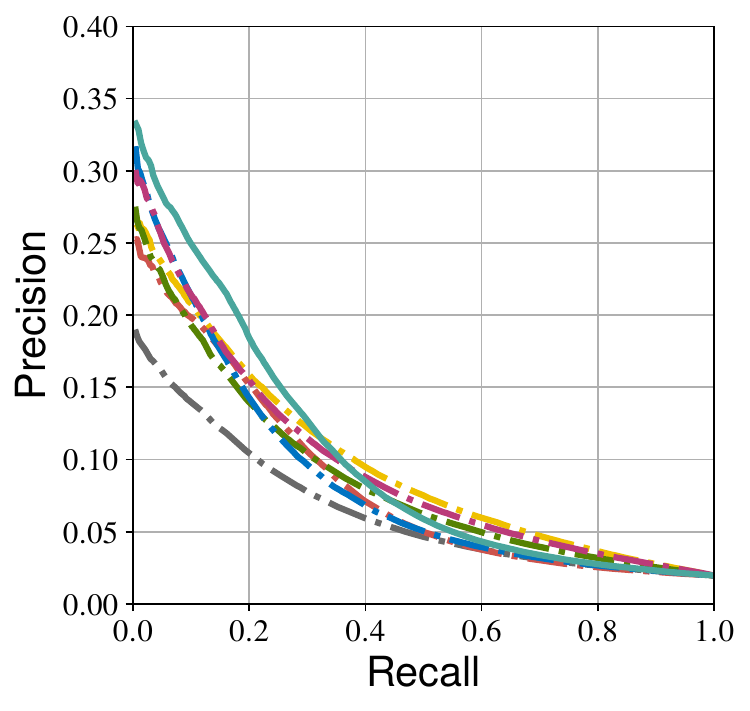}}
\caption{Retrieval PR curves of different models on the UCF101 and HMDB51 datasets.}
\label{fig:PR_curve}
\end{figure}

To investigate the retrieval performance in a wider range of ranking positions, we further present the PR curves of different models. 
As shown in \Cref{fig:PR_curve}, \modelname{} consistently achieves higher precision at the same recall rate across all code lengths compared to other state-of-the-art methods. 

\subsubsection{Cross-Dataset Transferability}
We evaluate the cross-dataset retrieval performance of different methods.
Specifically, we train them on UCF101 and test them on HMDB51. 
\Cref{tab:cross_dataset} presents the holistic results of different methods with 16-bit hash codes in the cross-dataset scenario. \modelname{} continues to outperform existing models, showcasing its exceptional ability to generalize across diverse video data.

\begin{table}[t]
\centering
\resizebox{\columnwidth}{!}{
    \begin{tabular}{ccccccc}
        \toprule
        SSVH & BTH & DKPH & MCMSH & ConMH & BerVAE & \textbf{\modelname{}} \\
        \midrule
        0.0187 & 0.0210 & 0.0243 & 0.0229 & 0.0285 & 0.0335 & \textbf{0.0378} \\
        \bottomrule
    \end{tabular}
}
\caption{Cross-dataset retrieval performance by GmAP. 
We train on UCF101 and test on HMDB51 using 16-bit models.}
\label{tab:cross_dataset}
\end{table}

\subsubsection{Inference Efficiency}
Inference efficiency is a crucial aspect of practical retrieval systems. 
Here we focus on inference time and memory overheads for producing video hash codes. 
We compare \modelname{} with 2 representative baselines, \ie the Transformer-based model ConMH, and the MCMSH based on MLP-Mixer~\cite{MLPMixer_2021}. 
We perform \emph{stress testing} with them in the same computational environment, taking 5 samples as a unit to probe the maximally affordable batchsizes and measuring the average inference time per sample. 
Since efficiency is independent of the retrieval performance, we directly stimulate tensors in various lengths as the testing input. 
The results are shown in \Cref{fig:intro}(b). 
It is clear that \modelname{} enjoys a larger inference throughput and faster processing speed. 
Moreover, \modelname{} exhibits ever-pronounced advantages with longer sequences. 

Based on the measured values, we use the least squares estimation to fit the scaling laws of inference time $T$ (in ms) \wrt input length $L$.
The functions of the 3 models are given by
$\hat{T}_\text{ConMH}=2.4\times10^{-6}L^2+1.9\times10^{-3}L+3.0\times10^{-2}$ ($R^2\approx0.999$); 
$\hat{T}_\text{MCMSH}=1.3\times10^{-6}L^2+2.1\times10^{-3}L+2.3\times10^{-2}$ ($R^2\approx0.999$);
$\hat{T}_\text{\modelname{}}=1.5\times10^{-3}L+3.4\times10^{-2}$ ($R^2\approx0.998$). 
We note that MLP-Mixer has linear complexity, but MCMSH focuses more on using its structure to boost performance, where the improved designs result in overall quadratic complexity. 
Different from ConMH and MCMSH, \modelname{} enjoys a preferable linear complexity. 

\subsection{Model Analyses}
\subsubsection{Effectiveness of Bidirectional Design}
We conduct an ablation study on the bidirectional design of \modelname{} and compare three block design strategies: (\textbf{i}) Forward Only, where the Mamba block processes the video sequence forward; (\textbf{ii}) Backward Only, where it processes the sequence backward; and (\textbf{iii}) Bidirectional (default), where stacked blocks process the sequence in both directions.
As shown in \Cref{tab:ssm_geomean}, the bidirectional design improves retrieval performance by 1\% to 6\% compared to both Variant (1) and Variant (2). This advantage comes from its ability to capture temporal dependencies in both directions. By leveraging information from both past and future frames, the Bidirectional design creates a more comprehensive representation of the video sequence.

\subsubsection{Effects of Different Loss Terms}
To analyse the effectiveness of three loss terms (\ie$\calL_\mathsf{TR}$ , $\calL_\mathsf{CL}$ and $\calL_\mathsf{CA}$) of \modelname{}, we construct several \modelname{} variants: 
(\textbf{i}) $\calL_\mathsf{TR}$ Only: train the model with merely temporal reconstruction. 
(\textbf{ii}) w/o $\calL_\mathsf{CA}$: \modelname{} removes the hash center alignment task.
(\textbf{iii}) w/o $\calL_\mathsf{CL}$: We train the model without contrastive learning. 
As shown in \Cref{tab:ssm_geomean}, the worst performance occurs when only $\calL_\mathsf{TR}$ is used. Comparing Variant (4) with Variant (3), adding $\calL_\mathsf{CL}$ significantly increases the GmAP, which can validate its necessity. Similarly, comparing Variant (5) with Variant (3)  and Figure \ref{fig:intro}(c), incorporating $\calL_\mathsf{CA}$ not only boosts GmAP accuracy by about 6\% but also accelerates convergence by approximately 65\%, underscoring the importance of $\calL_\mathsf{CA}$.

\subsubsection{Experiment with SSM Variants}
\begin{table}[t]
\centering
\resizebox{\columnwidth}{!}{
    \begin{tabular}{ccccccccc}
        \toprule
        \multirow{2}{*}{ID} & \multirow{2}{*}{Method} & \multicolumn{3}{c}{UCF101} && \multicolumn{3}{c}{HMDB51} \\
        \cmidrule{3-5} \cmidrule{7-9}
        && 16 bits & 32 bits & 64 bits && 16 bits & 32 bits & 64 bits \\
        \midrule
        (0) & \textbf{S5VH (full)}     & \textbf{0.357} & \textbf{0.390} & \textbf{0.440} && \textbf{0.093} & \textbf{0.130} & \textbf{0.142} \\
        \midrule\midrule
        (1) & Forward Only             & 0.351 & 0.375 & 0.424 && 0.079 & 0.123 & 0.136 \\
        (2) & Backward Only            & 0.324 & 0.337 & 0.390 && 0.070 & 0.120 & 0.133 \\
        \midrule\midrule
        (3) & $\calL_\mathsf{TR}$ Only & 0.138 & 0.210 & 0.280 && 0.045 & 0.075 & 0.081 \\
        (4) & w/o $\calL_\mathsf{CA}$  & 0.286 & 0.342 & 0.407 && 0.071 & 0.100 & 0.118 \\
        (5) & w/o $\calL_\mathsf{CL}$  & 0.204 & 0.217 & 0.290 && 0.070 & 0.095 & 0.102 \\
        \midrule\midrule
        (6) & w/ LSTM                  & 0.303 & 0.351 & 0.432 && 0.084 & 0.125 & 0.137 \\
        (7) & w/ RetNet                & 0.278 & 0.355 & 0.413 && 0.079 & 0.108 & 0.134 \\
        (8) & w/ RWKV                  & 0.307 & 0.340 & 0.408 && 0.074 & 0.106 & 0.135 \\
        \hline
        \bottomrule
    \end{tabular}
}
\caption{Ablation study of \modelname{}. We use the GmAP metric.}
\label{tab:ssm_geomean}
\end{table}

In addition to Mamba, other notable SSM architectures like RetNet \cite{RetNet_2023} and RWKV \cite{RWKV_2023} have shown excellent performance in various tasks. 
We design \modelname{} Variants (6)-(8), equipped with LSTM \cite{lstm} and the two architectures, to provide more insights for practical choices. 
To ensure a fair comparison, we only replace the Mamba layer with the corresponding bidirectional layers, while keeping all other factors the same. 

According to \Cref{tab:ssm_geomean}, we validate that the default choice of Mamba is satisfactory to show the best performance. 
We notice that LSTM is still a strong option in SSVH, even though it has become less popular in recent years. 

\begin{figure}[t]
\centering
\setcounter{subfigure}{0}
\subfloat[\small{ConMH 16 bits}]{\includegraphics[width = 0.1600\textwidth]{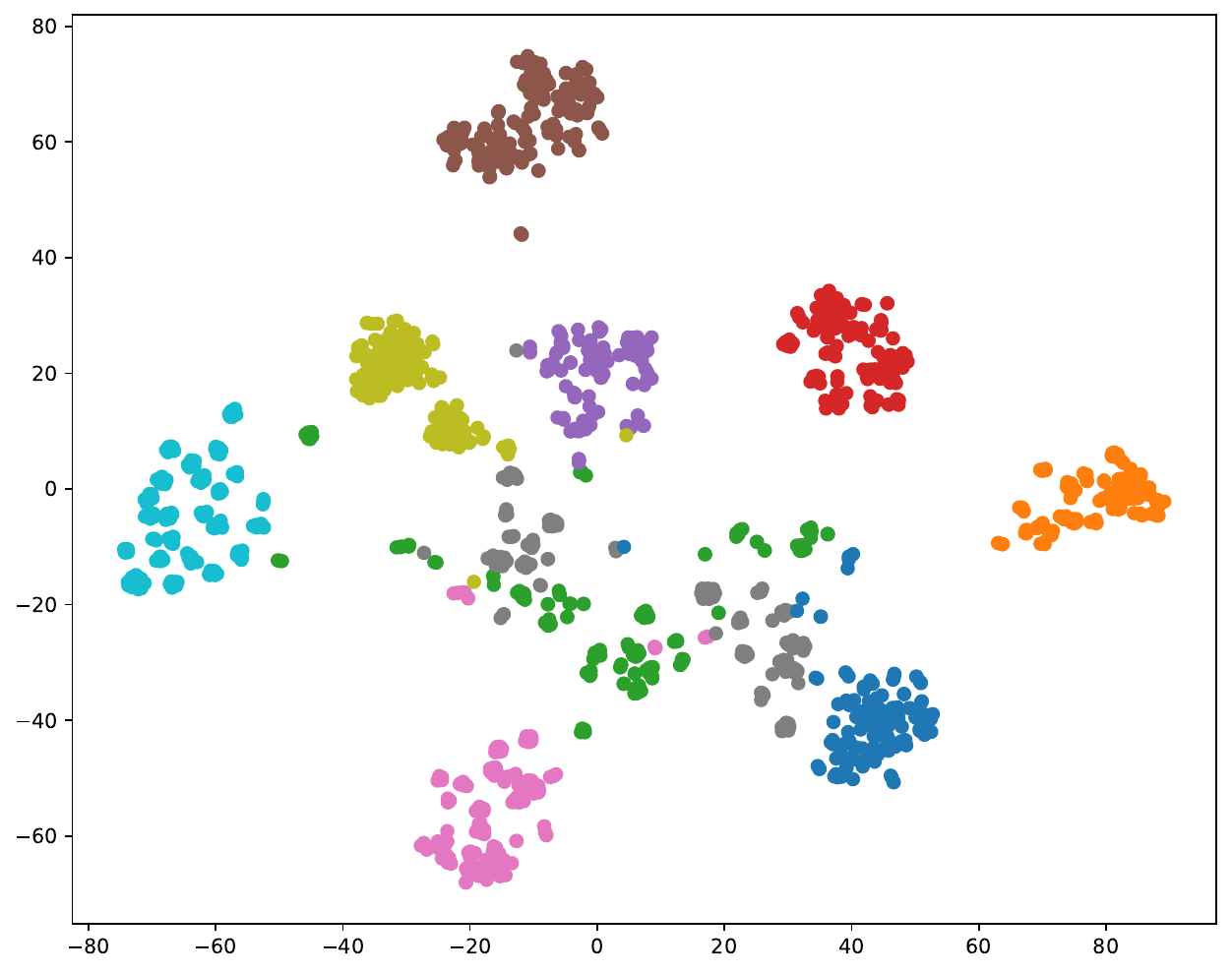}}
\subfloat[\small{ConMH 32 bits}]{\includegraphics[width = 0.1600\textwidth]{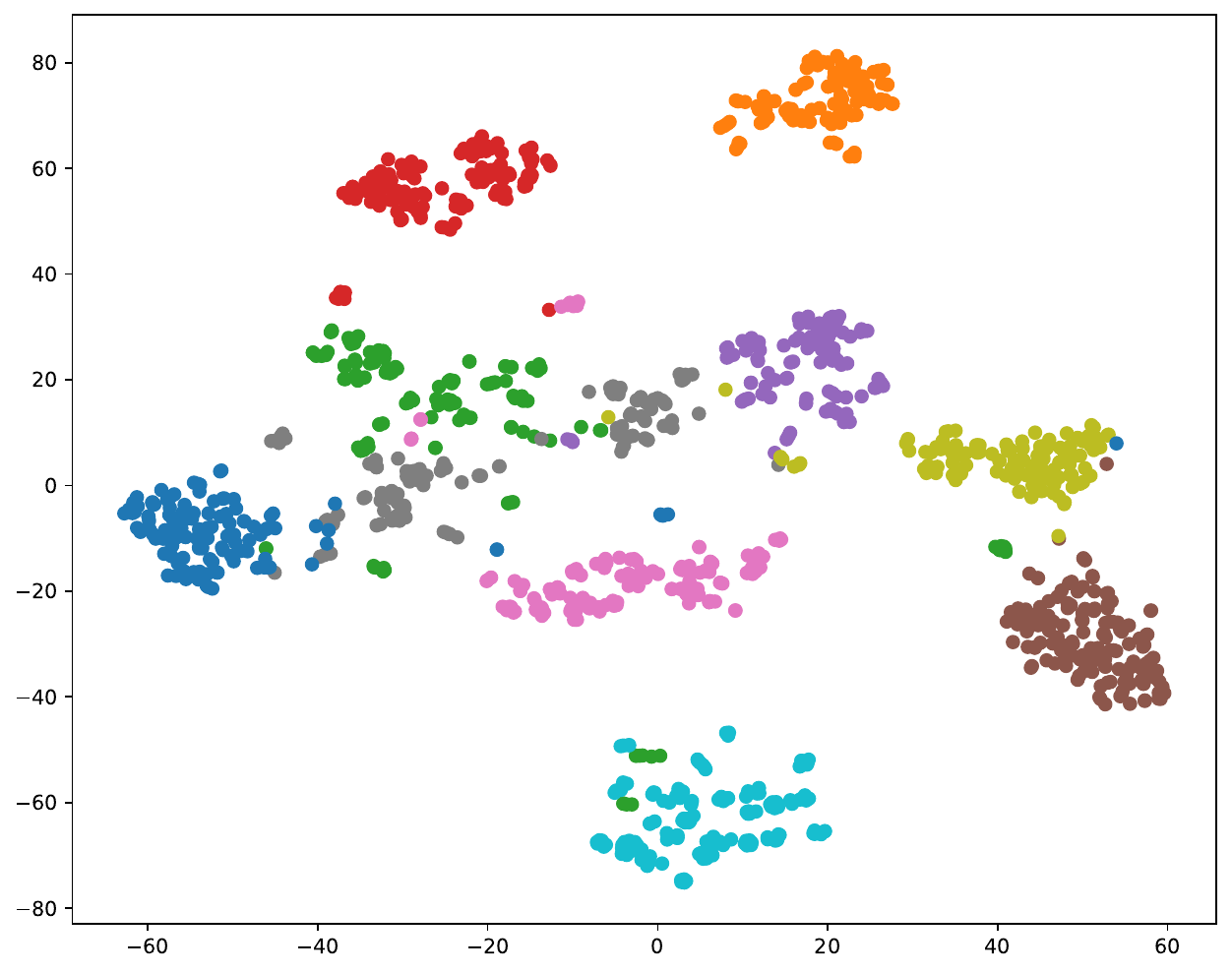}}
\subfloat[\small{ConMH 64 bits}]{\includegraphics[width = 0.1600\textwidth]{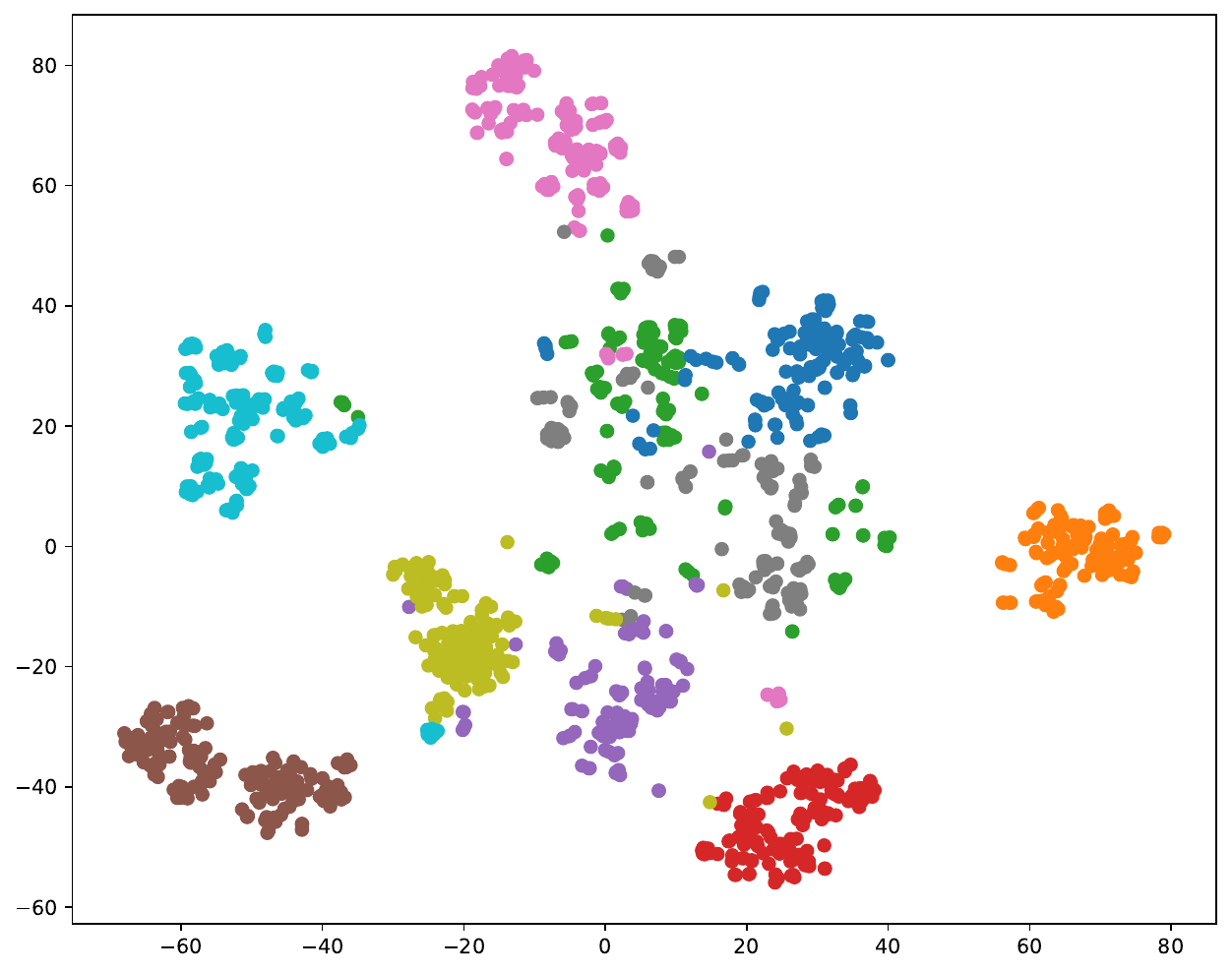}}

\subfloat[\small{\modelname{} 16 bits}]{\includegraphics[width = 0.1600\textwidth]{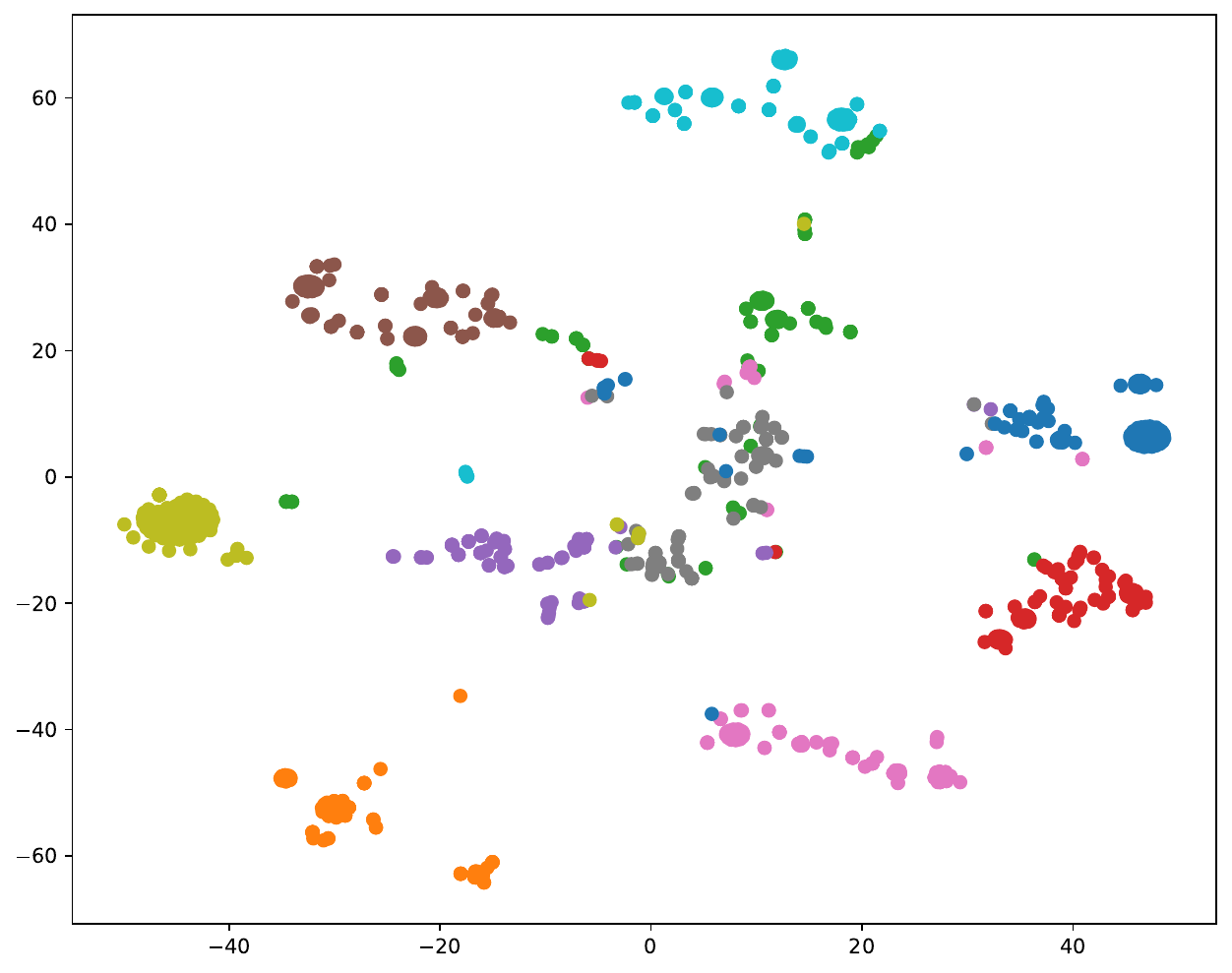}}
\subfloat[\small{\modelname{} 32 bits}]{\includegraphics[width = 0.1600\textwidth]{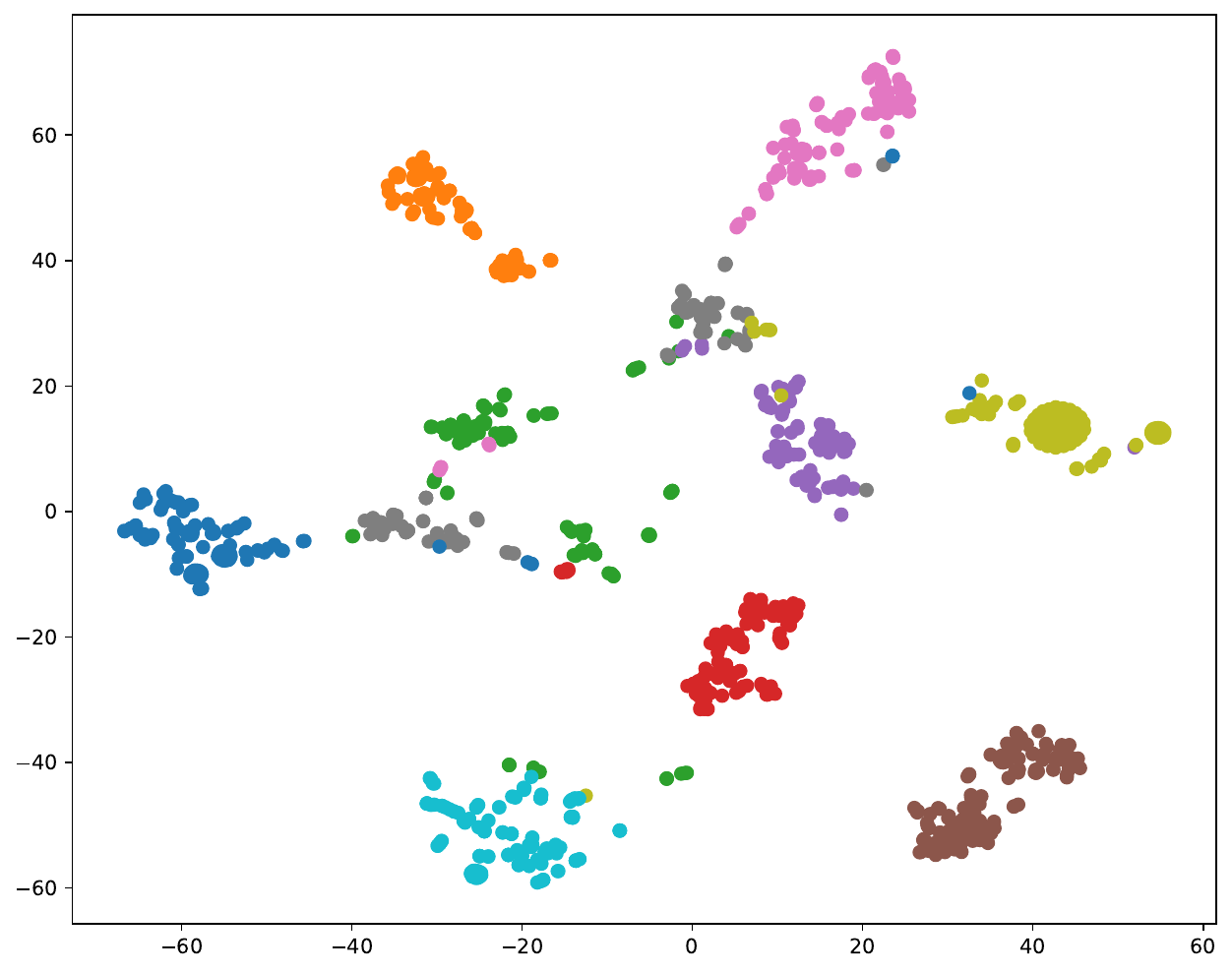}}
\subfloat[\small{\modelname{} 64 bits}]{\includegraphics[width = 0.1600\textwidth]{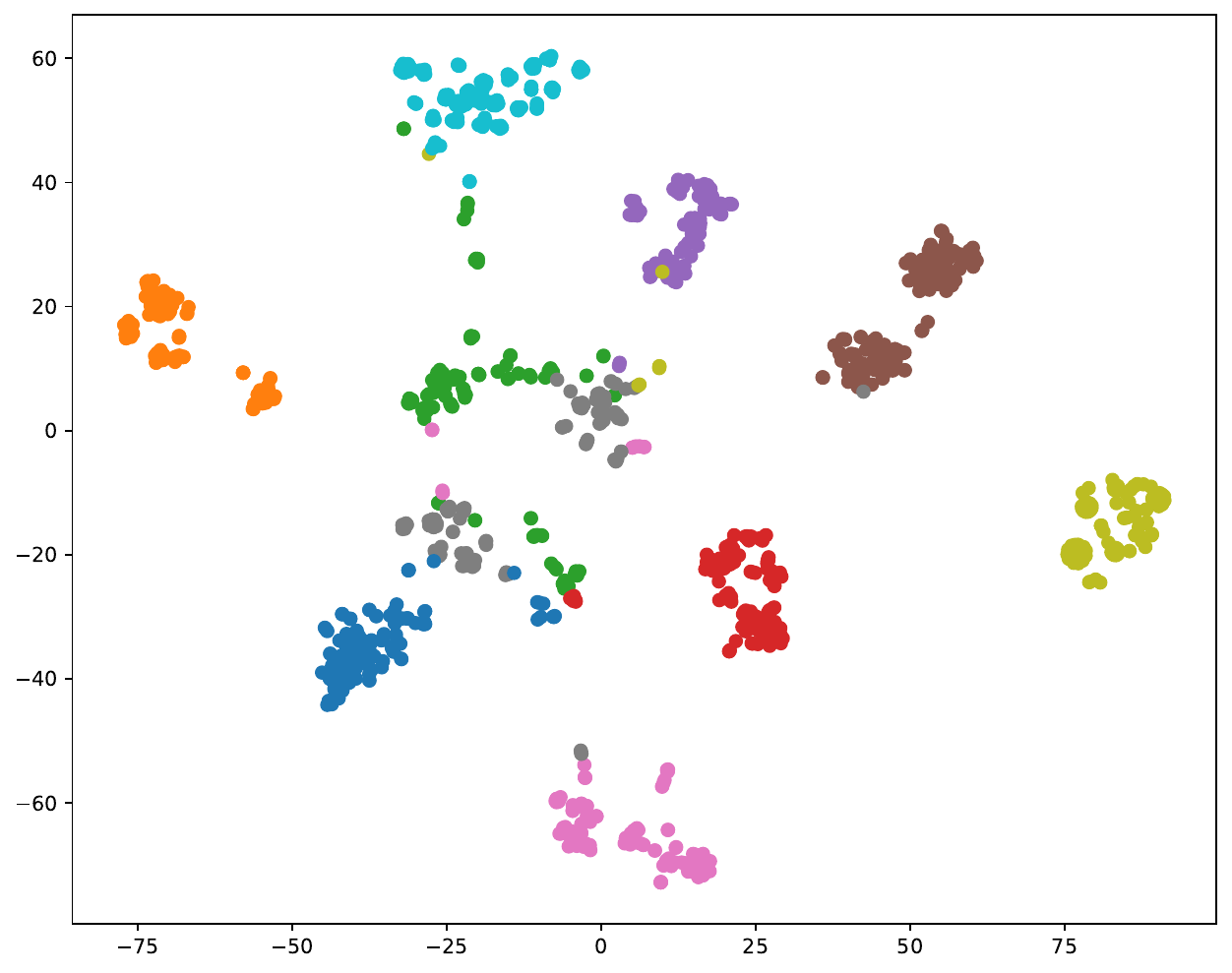}}
\caption{The t-SNE visualization of the learned hash codes on UCF101. Data points of the same color correspond to the same category. Only the first 10 classes are visualized.}
\label{fig:TSNE}
\end{figure}

\subsection{Qualitative Results}
To further analyze the results, we utilize t-SNE \cite{tsne2008JMLR} to visualize the learned hash codes on the UCF101 dataset. 
As illustrated in \Cref{fig:TSNE}, compared to the generated hash codes of ConMH, the hash codes generated by \modelname{} demonstrate clearer compactness within the same category and increased separation between different categories. This finding indicates that \modelname{} produces more discriminative binary codes, which significantly improves retrieval performance.

%% file: sections/Conclusions.tex
\section{Conclusions}
\label{sec:conclusion}
In this paper, we introduced \modelname{}, the first Mamba-based SSVH model with an enhanced learning paradigm. 
\modelname{} develop bidirectional Mamba layers to capture comprehensive temporal relations for hash learning. 
To improve training efficiency, we proposed a semantic hash center generation algorithm and a center alignment loss to extract and leverage the global learning signal. 
Experiments show \modelname{}'s consistent improvements under various setups, transfers better, and superior inference efficiency. 
Our study suggests the strong potential of state-space models in video hashing, which we hope can inspire further research.

\section*{Acknowledgements}
This work is supported in part by the National Natural Science Foundation of China under grant 624B2088, 62171248, 62301189, and Shenzhen Science and Technology Program under Grant JCYJ20220818101012025, RCBS20221008093124061, GXWD20220811172936001.

%% file: main.bbl
\begin{thebibliography}{58}
\providecommand{\natexlab}[1]{#1}

\bibitem[{Bengio, L{\'e}onard, and Courville(2013)}]{STE_2013}
Bengio, Y.; L{\'e}onard, N.; and Courville, A. 2013.
\newblock Estimating or propagating gradients through stochastic neurons for conditional computation.
\newblock \emph{arXiv preprint arXiv:1308.3432}.

\bibitem[{Caba~Heilbron et~al.(2015)Caba~Heilbron, Escorcia, Ghanem, and Carlos~Niebles}]{caba2015activitynet}
Caba~Heilbron, F.; Escorcia, V.; Ghanem, B.; and Carlos~Niebles, J. 2015.
\newblock Activitynet: A large-scale video benchmark for human activity understanding.
\newblock In \emph{Proceedings of the ieee conference on computer vision and pattern recognition}, 961--970.

\bibitem[{Duan et~al.(2024)Duan, Hao, Zhu, Cheng, Zhou, and Wang}]{EUVH_2024}
Duan, J.; Hao, Y.; Zhu, B.; Cheng, L.; Zhou, P.; and Wang, X. 2024.
\newblock Efficient Unsupervised Video Hashing with Contextual Modeling and Structural Controlling.
\newblock \emph{IEEE Transactions on Multimedia}.

\bibitem[{Fu et~al.(2022)Fu, Dao, Saab, Thomas, Rudra, and R{\'e}}]{fu2022hungry}
Fu, D.~Y.; Dao, T.; Saab, K.~K.; Thomas, A.~W.; Rudra, A.; and R{\'e}, C. 2022.
\newblock Hungry hungry hippos: Towards language modeling with state space models.
\newblock \emph{arXiv preprint arXiv:2212.14052}.

\bibitem[{Gao et~al.(2023)Gao, Bai, Chen, Wu, and Xia}]{gao2023backdoor}
Gao, K.; Bai, J.; Chen, B.; Wu, D.; and Xia, S.-T. 2023.
\newblock Backdoor Attack on Hash-based Image Retrieval via Clean-label Data Poisoning.
\newblock In \emph{The 34th British Machine Vision Conference}, 172--173.

\bibitem[{Gong et~al.(2012)Gong, Lazebnik, Gordo, and Perronnin}]{ITQ_2012}
Gong, Y.; Lazebnik, S.; Gordo, A.; and Perronnin, F. 2012.
\newblock Iterative quantization: A procrustean approach to learning binary codes for large-scale image retrieval.
\newblock \emph{IEEE transactions on pattern analysis and machine intelligence}, 35(12): 2916--2929.

\bibitem[{Gu and Dao(2023)}]{Mamba_2023}
Gu, A.; and Dao, T. 2023.
\newblock Mamba: Linear-time sequence modeling with selective state spaces.
\newblock \emph{arXiv preprint arXiv:2312.00752}.

\bibitem[{Gu et~al.(2020)Gu, Dao, Ermon, Rudra, and R{\'e}}]{gu2020hippo}
Gu, A.; Dao, T.; Ermon, S.; Rudra, A.; and R{\'e}, C. 2020.
\newblock Hippo: Recurrent memory with optimal polynomial projections.
\newblock \emph{Advances in Neural Information Processing Systems}, 33.

\bibitem[{Gu et~al.(2022)Gu, Goel, Gupta, and R{\'e}}]{S4D_2022}
Gu, A.; Goel, K.; Gupta, A.; and R{\'e}, C. 2022.
\newblock On the parameterization and initialization of diagonal state space models.
\newblock \emph{Advances in Neural Information Processing Systems}, 35: 35971--35983.

\bibitem[{Gu, Goel, and R{\'e}(2021)}]{S4_2021}
Gu, A.; Goel, K.; and R{\'e}, C. 2021.
\newblock Efficiently Modeling Long Sequences with Structured State Spaces.
\newblock In \emph{International Conference on Learning Representations}.

\bibitem[{Gu et~al.(2021)Gu, Johnson, Goel, Saab, Dao, Rudra, and R{\'e}}]{LSSL_2021}
Gu, A.; Johnson, I.; Goel, K.; Saab, K.; Dao, T.; Rudra, A.; and R{\'e}, C. 2021.
\newblock Combining recurrent, convolutional, and continuous-time models with linear state space layers.
\newblock \emph{Advances in neural information processing systems}, 34: 572--585.

\bibitem[{Gupta, Gu, and Berant(2022)}]{DSS_2022}
Gupta, A.; Gu, A.; and Berant, J. 2022.
\newblock Diagonal state spaces are as effective as structured state spaces.
\newblock \emph{Advances in Neural Information Processing Systems}, 35: 22982--22994.

\bibitem[{Hao et~al.(2022)Hao, Duan, Zhang, Zhu, Zhou, and He}]{MCMSH_2022}
Hao, Y.; Duan, J.; Zhang, H.; Zhu, B.; Zhou, P.; and He, X. 2022.
\newblock Unsupervised video hashing with multi-granularity contextualization and multi-structure preservation.
\newblock In \emph{Proceedings of the 30th ACM International Conference on Multimedia}, 3754--3763.

\bibitem[{Hao et~al.(2017)Hao, Mu, Goulermas, Jiang, Hong, and Wang}]{tUSMVH_2017}
Hao, Y.; Mu, T.; Goulermas, J.~Y.; Jiang, J.; Hong, R.; and Wang, M. 2017.
\newblock Unsupervised t-distributed video hashing and its deep hashing extension.
\newblock \emph{IEEE Transactions on Image Processing}, 26(11): 5531--5544.

\bibitem[{He et~al.(2020)He, Fan, Wu, Xie, and Girshick}]{MoCo_2020}
He, K.; Fan, H.; Wu, Y.; Xie, S.; and Girshick, R. 2020.
\newblock Momentum contrast for unsupervised visual representation learning.
\newblock In \emph{Proceedings of the IEEE/CVF conference on computer vision and pattern recognition}, 9729--9738.

\bibitem[{He et~al.(2016)He, Zhang, Ren, and Sun}]{he2016deep}
He, K.; Zhang, X.; Ren, S.; and Sun, J. 2016.
\newblock Deep residual learning for image recognition.
\newblock In \emph{Proceedings of the IEEE conference on computer vision and pattern recognition}, 770--778.

\bibitem[{Hochreiter and Schmidhuber(1997)}]{lstm}
Hochreiter, S.; and Schmidhuber, J. 1997.
\newblock Long Short-Term Memory.
\newblock \emph{Neural Computation}, 9(8): 1735--1780.

\bibitem[{Jiang et~al.(2017)Jiang, Wu, Wang, Xue, and Chang}]{jiang2017exploiting}
Jiang, Y.-G.; Wu, Z.; Wang, J.; Xue, X.; and Chang, S.-F. 2017.
\newblock Exploiting feature and class relationships in video categorization with regularized deep neural networks.
\newblock \emph{IEEE transactions on pattern analysis and machine intelligence}, 40(2): 352--364.

\bibitem[{Kalman(1960)}]{kalman1960new}
Kalman, R.~E. 1960.
\newblock {A New Approach to Linear Filtering and Prediction Problems}.
\newblock \emph{Journal of Basic Engineering}, 82(1): 35--45.

\bibitem[{Kuehne et~al.(2011)Kuehne, Jhuang, Garrote, Poggio, and Serre}]{HMDB2011ICCV}
Kuehne, H.; Jhuang, H.; Garrote, E.; Poggio, T.~A.; and Serre, T. 2011.
\newblock {HMDB:} {A} large video database for human motion recognition.
\newblock In Metaxas, D.~N.; Quan, L.; Sanfeliu, A.; and Gool, L.~V., eds., \emph{{IEEE} International Conference on Computer Vision, {ICCV} 2011, Barcelona, Spain, November 6-13, 2011}, 2556--2563. {IEEE} Computer Society.

\bibitem[{Li et~al.(2017)Li, Yang, Cao, and Huang}]{JTAE_2017}
Li, C.; Yang, Y.; Cao, J.; and Huang, Z. 2017.
\newblock Jointly modeling static visual appearance and temporal pattern for unsupervised video hashing.
\newblock In \emph{Proceedings of the 2017 ACM on Conference on Information and Knowledge Management}, 9--17.

\bibitem[{Li et~al.(2022)Li, Xie, Ge, Zhang, Min, and Zhang}]{DKPH_2022}
Li, P.; Xie, H.; Ge, J.; Zhang, L.; Min, S.; and Zhang, Y. 2022.
\newblock Dual-stream knowledge-preserving hashing for unsupervised video retrieval.
\newblock In \emph{European Conference on Computer Vision}, 181--197. Springer.

\bibitem[{Li, Tian, and Ng(2024)}]{TSVH_2024}
Li, Q.; Tian, X.; and Ng, W.~W. 2024.
\newblock Self-supervised Temporal Sensitive Hashing for Video Retrieval.
\newblock \emph{IEEE Transactions on Multimedia}.

\bibitem[{Li et~al.(2019{\natexlab{a}})Li, Chen, Li, Lu, and Zhou}]{UVVH_2019}
Li, S.; Chen, Z.; Li, X.; Lu, J.; and Zhou, J. 2019{\natexlab{a}}.
\newblock Unsupervised variational video hashing with 1D-CNN-LSTM networks.
\newblock \emph{IEEE Transactions on Multimedia}, 22(6): 1542--1554.

\bibitem[{Li et~al.(2019{\natexlab{b}})Li, Chen, Lu, Li, and Zhou}]{NPH_2019}
Li, S.; Chen, Z.; Lu, J.; Li, X.; and Zhou, J. 2019{\natexlab{b}}.
\newblock Neighborhood preserving hashing for scalable video retrieval.
\newblock In \emph{Proceedings of the IEEE/CVF International Conference on Computer Vision}, 8212--8221.

\bibitem[{Li et~al.(2021)Li, Li, Lu, and Zhou}]{BTH_2021}
Li, S.; Li, X.; Lu, J.; and Zhou, J. 2021.
\newblock Self-supervised video hashing via bidirectional transformers.
\newblock In \emph{Proceedings of the IEEE/CVF Conference on Computer Vision and Pattern Recognition}, 13549--13558.

\bibitem[{Liu et~al.(2024)Liu, Tian, Zhao, Yu, Xie, Wang, Ye, and Liu}]{VMamba_2024}
Liu, Y.; Tian, Y.; Zhao, Y.; Yu, H.; Xie, L.; Wang, Y.; Ye, Q.; and Liu, Y. 2024.
\newblock Vmamba: Visual state space model.
\newblock \emph{arXiv preprint arXiv:2401.10166}.

\bibitem[{Peng et~al.(2023)Peng, Alcaide, Anthony, Albalak, Arcadinho, Biderman, Cao, Cheng, Chung, Derczynski et~al.}]{RWKV_2023}
Peng, B.; Alcaide, E.; Anthony, Q.; Albalak, A.; Arcadinho, S.; Biderman, S.; Cao, H.; Cheng, X.; Chung, M.; Derczynski, L.; et~al. 2023.
\newblock RWKV: Reinventing RNNs for the Transformer Era.
\newblock In \emph{Findings of the Association for Computational Linguistics: EMNLP 2023}, 14048--14077.

\bibitem[{Qiao et~al.(2024)Qiao, Yu, Guo, Chen, Zhao, Sun, Wu, and Liu}]{VLMamba_2024}
Qiao, Y.; Yu, Z.; Guo, L.; Chen, S.; Zhao, Z.; Sun, M.; Wu, Q.; and Liu, J. 2024.
\newblock Vl-mamba: Exploring state space models for multimodal learning.
\newblock \emph{arXiv preprint arXiv:2403.13600}.

\bibitem[{Ramachandran, Zoph, and Le(2017)}]{SiLU_2017}
Ramachandran, P.; Zoph, B.; and Le, Q.~V. 2017.
\newblock Searching for activation functions.
\newblock \emph{arXiv preprint arXiv:1710.05941}.

\bibitem[{Shen et~al.(2023)Shen, Zhou, Yuan, Yang, Lan, and Zheng}]{CTH_2023}
Shen, X.; Zhou, Y.; Yuan, Y.-H.; Yang, X.; Lan, L.; and Zheng, Y. 2023.
\newblock Contrastive Transformer Hashing for Compact Video Representation.
\newblock \emph{IEEE Transactions on Image Processing}.

\bibitem[{Simonyan and Zisserman(2014)}]{simonyan2014very}
Simonyan, K.; and Zisserman, A. 2014.
\newblock Very deep convolutional networks for large-scale image recognition.
\newblock \emph{arXiv preprint arXiv:1409.1556}.

\bibitem[{Smith, Warrington, and Linderman(2022)}]{smith2022simplified}
Smith, J.~T.; Warrington, A.; and Linderman, S.~W. 2022.
\newblock Simplified state space layers for sequence modeling.
\newblock \emph{arXiv preprint arXiv:2208.04933}.

\bibitem[{Song et~al.(2011)Song, Yang, Huang, Shen, and Hong}]{MFH_2011}
Song, J.; Yang, Y.; Huang, Z.; Shen, H.~T.; and Hong, R. 2011.
\newblock Multiple feature hashing for real-time large scale near-duplicate video retrieval.
\newblock In \emph{Proceedings of the 19th ACM international conference on Multimedia}, 423--432.

\bibitem[{Song et~al.(2018{\natexlab{a}})Song, Zhang, Li, Gao, Wang, and Hong}]{SSVH_2018}
Song, J.; Zhang, H.; Li, X.; Gao, L.; Wang, M.; and Hong, R. 2018{\natexlab{a}}.
\newblock Self-supervised video hashing with hierarchical binary auto-encoder.
\newblock \emph{IEEE Transactions on Image Processing}, 27(7): 3210--3221.

\bibitem[{Song et~al.(2018{\natexlab{b}})Song, Zhang, Li, Gao, Wang, and Hong}]{song2018self}
Song, J.; Zhang, H.; Li, X.; Gao, L.; Wang, M.; and Hong, R. 2018{\natexlab{b}}.
\newblock Self-supervised video hashing with hierarchical binary auto-encoder.
\newblock \emph{IEEE Transactions on Image Processing}, 27(7): 3210--3221.

\bibitem[{Soomro, Zamir, and Shah(2012)}]{UCF-1012012arxiv}
Soomro, K.; Zamir, A.~R.; and Shah, M. 2012.
\newblock {UCF101:} {A} Dataset of 101 Human Actions Classes From Videos in The Wild.
\newblock \emph{CoRR}, abs/1212.0402.

\bibitem[{Sun et~al.(2023{\natexlab{a}})Sun, Dong, Huang, Ma, Xia, Xue, Wang, and Wei}]{RetNet_2023}
Sun, Y.; Dong, L.; Huang, S.; Ma, S.; Xia, Y.; Xue, J.; Wang, J.; and Wei, F. 2023{\natexlab{a}}.
\newblock Retentive network: A successor to transformer for large language models.
\newblock \emph{arXiv preprint arXiv:2307.08621}.

\bibitem[{Sun et~al.(2024)Sun, Qin, Peng, Ren, Yang, and Hu}]{sun2024dual}
Sun, Y.; Qin, Y.; Peng, D.; Ren, Z.; Yang, C.; and Hu, P. 2024.
\newblock Dual Self-Paced Hashing for Image Retrieval.
\newblock \emph{IEEE Transactions on Multimedia}, 26: 9619--9629.

\bibitem[{Sun et~al.(2023{\natexlab{b}})Sun, Ren, Hu, Peng, and Wang}]{sun2023hierarchical}
Sun, Y.; Ren, Z.; Hu, P.; Peng, D.; and Wang, X. 2023{\natexlab{b}}.
\newblock Hierarchical consensus hashing for cross-modal retrieval.
\newblock \emph{IEEE Transactions on Multimedia}, 26: 824--836.

\bibitem[{Tolstikhin et~al.(2021)Tolstikhin, Houlsby, Kolesnikov, Beyer, Zhai, Unterthiner, Yung, Steiner, Keysers, Uszkoreit et~al.}]{MLPMixer_2021}
Tolstikhin, I.~O.; Houlsby, N.; Kolesnikov, A.; Beyer, L.; Zhai, X.; Unterthiner, T.; Yung, J.; Steiner, A.; Keysers, D.; Uszkoreit, J.; et~al. 2021.
\newblock Mlp-mixer: An all-mlp architecture for vision.
\newblock \emph{Advances in neural information processing systems}, 34: 24261--24272.

\bibitem[{van~der Maaten and Hinton(2008)}]{tsne2008JMLR}
van~der Maaten, L.; and Hinton, G.~E. 2008.
\newblock Visualizing Data using t-SNE.
\newblock \emph{Journal of Machine Learning Research}, 9: 2579--2605.

\bibitem[{Vaswani et~al.(2017)Vaswani, Shazeer, Parmar, Uszkoreit, Jones, Gomez, Kaiser, and Polosukhin}]{vaswani2017attention}
Vaswani, A.; Shazeer, N.; Parmar, N.; Uszkoreit, J.; Jones, L.; Gomez, A.~N.; Kaiser, {\L}.; and Polosukhin, I. 2017.
\newblock Attention is all you need.
\newblock \emph{Advances in neural information processing systems}, 30.

\bibitem[{Veli{\v{c}}kovi{\'c} et~al.(2018)Veli{\v{c}}kovi{\'c}, Cucurull, Casanova, Romero, Li{\`o}, and Bengio}]{GAT_2018}
Veli{\v{c}}kovi{\'c}, P.; Cucurull, G.; Casanova, A.; Romero, A.; Li{\`o}, P.; and Bengio, Y. 2018.
\newblock Graph Attention Networks.
\newblock In \emph{International Conference on Learning Representations}.

\bibitem[{Wang et~al.(2024{\natexlab{a}})Wang, Tsepa, Ma, and Wang}]{GraphMamba_2024}
Wang, C.; Tsepa, O.; Ma, J.; and Wang, B. 2024{\natexlab{a}}.
\newblock Graph-mamba: Towards long-range graph sequence modeling with selective state spaces.
\newblock \emph{arXiv preprint arXiv:2402.00789}.

\bibitem[{Wang et~al.(2024{\natexlab{b}})Wang, Zeng, Chen, Wang, Liao, Li, Wang, and Xia}]{HuggingHash2_2024}
Wang, J.; Zeng, Z.; Chen, B.; Wang, Y.; Liao, D.; Li, G.; Wang, Y.; and Xia, S.-T. 2024{\natexlab{b}}.
\newblock Hugs Bring Double Benefits: Unsupervised Cross-Modal Hashing with Multi-granularity Aligned Transformers.
\newblock \emph{International Journal of Computer Vision}, 1--33.

\bibitem[{Wang et~al.(2023{\natexlab{a}})Wang, Wang, Chen, Zeng, and Xia}]{ConMH_2023}
Wang, Y.; Wang, J.; Chen, B.; Zeng, Z.; and Xia, S.-T. 2023{\natexlab{a}}.
\newblock Contrastive masked autoencoders for self-supervised video hashing.
\newblock In \emph{Proceedings of the AAAI Conference on Artificial Intelligence}, volume 37(3), 2733--2741.

\bibitem[{Wang et~al.(2023{\natexlab{b}})Wang, Zhou, Sun, and Qian}]{BerVAE_2023}
Wang, Y.; Zhou, M.; Sun, Y.; and Qian, X. 2023{\natexlab{b}}.
\newblock Uncertainty-aware unsupervised video hashing.
\newblock In \emph{International Conference on Artificial Intelligence and Statistics}, 6722--6740.

\bibitem[{Wei et~al.(2023)Wei, Liu, Song, Cui, Xie, and Zhou}]{CHAIN_2023}
Wei, R.; Liu, Y.; Song, J.; Cui, H.; Xie, Y.; and Zhou, K. 2023.
\newblock CHAIN: Exploring Global-Local Spatio-Temporal Information for Improved Self-Supervised Video Hashing.
\newblock In \emph{Proceedings of the 31st ACM International Conference on Multimedia}, 1677--1688.

\bibitem[{Weiss, Torralba, and Fergus(2008)}]{SH_2008}
Weiss, Y.; Torralba, A.; and Fergus, R. 2008.
\newblock Spectral hashing.
\newblock \emph{Advances in neural information processing systems}, 21.

\bibitem[{Wu and Ghanem(2018)}]{LpBoxADMM_2018}
Wu, B.; and Ghanem, B. 2018.
\newblock $\ell_p$-Box ADMM: A Versatile Framework for Integer Programming.
\newblock \emph{IEEE transactions on pattern analysis and machine intelligence}, 41(7): 1695--1708.

\bibitem[{Wu et~al.(2017)Wu, Liu, Guo, Ding, Han, Shen, and Shao}]{wu2017unsupervised}
Wu, G.; Liu, L.; Guo, Y.; Ding, G.; Han, J.; Shen, J.; and Shao, L. 2017.
\newblock Unsupervised deep video hashing with balanced rotation.
\newblock In \emph{Proceedings of the 26th International Joint Conference on Artificial Intelligence}, 3076--3082.

\bibitem[{Ye et~al.(2013)Ye, Liu, Wang, and Chang}]{VHDT_2013}
Ye, G.; Liu, D.; Wang, J.; and Chang, S.-F. 2013.
\newblock Large-scale video hashing via structure learning.
\newblock In \emph{Proceedings of the IEEE international conference on computer vision}, 2272--2279.

\bibitem[{Zeng et~al.(2022)Zeng, Wang, Chen, Wang, and Xia}]{MAGRH_2022}
Zeng, Z.; Wang, J.; Chen, B.; Wang, Y.; and Xia, S.-T. 2022.
\newblock Motion-Aware Graph Reasoning Hashing for Self-supervised Video Retrieval.
\newblock In \emph{33rd British Machine Vision Conference}, 82.

\bibitem[{Zhang et~al.(2016)Zhang, Wang, Hong, and Chua}]{SSTH_2016}
Zhang, H.; Wang, M.; Hong, R.; and Chua, T.-S. 2016.
\newblock Play and rewind: Optimizing binary representations of videos by self-supervised temporal hashing.
\newblock In \emph{Proceedings of the 24th ACM international conference on Multimedia}, 781--790.

\bibitem[{Zhao et~al.(2024)Zhao, Zhang, Zhao, Ding, Huang, and Wang}]{Cobra_2024}
Zhao, H.; Zhang, M.; Zhao, W.; Ding, P.; Huang, S.; and Wang, D. 2024.
\newblock Cobra: Extending mamba to multi-modal large language model for efficient inference.
\newblock \emph{arXiv preprint arXiv:2403.14520}.

\bibitem[{Zhou et~al.(2024)Zhou, Sun, Liu, Chen, and Zhang}]{AVHash_2024}
Zhou, Y.; Sun, Z.; Liu, R.; Chen, Y.; and Zhang, D. 2024.
\newblock AVHash: Joint Audio-Visual Hashing for Video Retrieval.
\newblock In \emph{Proceedings of the 32nd ACM international conference on Multimedia}.

\bibitem[{Zhu et~al.(2024)Zhu, Liao, Zhang, Wang, Liu, and Wang}]{Vim_2024}
Zhu, L.; Liao, B.; Zhang, Q.; Wang, X.; Liu, W.; and Wang, X. 2024.
\newblock Vision mamba: Efficient visual representation learning with bidirectional state space model.
\newblock \emph{arXiv preprint arXiv:2401.09417}.

\end{thebibliography}
